\documentclass[letterpaper]{article} 
\usepackage{aaai2026}  
\usepackage{times}  
\usepackage{helvet}  
\usepackage{courier}  
\usepackage[hyphens]{url}  
\usepackage{graphicx} 
\usepackage{natbib}  
\usepackage{caption} 
\usepackage{algorithm}

\usepackage{newfloat}
\usepackage{listings}
\DeclareCaptionStyle{ruled}{labelfont=normalfont,labelsep=colon,strut=off} 
\floatstyle{ruled}
\newfloat{listing}{tb}{lst}{}
\floatname{listing}{Listing}
\usepackage{times}
\usepackage{latexsym}
\usepackage{mathtools}
\usepackage{amssymb}
\usepackage{multirow}
\usepackage[T1]{fontenc}

\usepackage[utf8]{inputenc}

\usepackage{microtype}

\usepackage{inconsolata}

\usepackage{colortbl}
\usepackage{booktabs}
\usepackage{amsmath}
\usepackage{algpseudocode}

\usepackage{soul}
\usepackage[dvipsnames]{xcolor}

\title{AlignTree: Efficient Defense Against LLM Jailbreak Attacks}
\author{
    Gil Goren, ~~
    Shahar Katz, ~~
    Lior Wolf
}
\affiliations{
    Blavatnik School of Computer Science, Tel Aviv University\\
    \{gilgoren@mail,shaharkatz3@mail,wolf@cs\}.tau.ac.il
}

\begin{document}

\maketitle

\begin{abstract}
Large Language Models (LLMs) are vulnerable to adversarial attacks that bypass safety guidelines and generate harmful content. Mitigating these vulnerabilities requires defense mechanisms that are both robust and computationally efficient. However, existing approaches either incur high computational costs or rely on lightweight defenses that can be easily circumvented, rendering them impractical for real-world LLM-based systems. In this work, we introduce the AlignTree defense, which enhances model alignment while maintaining minimal computational overhead. AlignTree monitors LLM activations during generation and detects misaligned behavior using an efficient random forest classifier. This classifier operates on two signals: (i) the refusal direction—a linear representation that activates on misaligned prompts, and (ii) an SVM-based signal that captures non-linear features associated with harmful content. Unlike previous methods, AlignTree does not require additional prompts or auxiliary guard models. Through extensive experiments, we demonstrate the efficiency and robustness of AlignTree across multiple LLMs and benchmarks.
Our code is available at:
\begin{links}
\link{Code}{https://github.com/Gilgo2/AlignTree}
\end{links}
\end{abstract}

\begin{table*}[t!]
\centering
\begin{small}
\begin{tabular}{lccc}
\toprule
 \multirow{2}{*}{Method} & \multirow{2}{*}{ASR $\downarrow$} & \multicolumn{2}{c}{Overhead} \\
                      &  & Additional LLM & Additional Inference \\
\midrule
Baseline model          & High & No  & 0   \\
Llama Guard \cite{meta2024llamaguard}   & Low          & Yes & 2  \\
AutoDefense \cite{zeng2024autodefensemultiagentllmdefense}         & Low   & Yes & 20 \\
SmoothLLM \cite{robey2024smoothllmdefendinglargelanguage}        & Medium      & No  & 10 \\
SelfDefense  \cite{phute2024llmselfdefenseself}        & Medium   & No  & 2  \\
PerplexityDefense \cite{jain2023baselinedefensesadversarialattacks}       & High & No  & 0   \\
AlignTree (\textbf{Ours})   & Low     & No  & 0   \\
\bottomrule
\end{tabular}
\end{small}
\caption{LLMs jailbreak defense methods and their computational overheads.}
\label{Summary overhead}
\end{table*}
\section{Introduction}
LLMs have become integral to numerous applications across various domains, making their security a pressing concern. However, recent research has highlighted vulnerabilities such as using LLMs to generate phishing emails, malicious code, hate speech, and inadvertently exposing sensitive information \cite{wei2023jailbrokendoesllmsafety, gupta2023chatgptthreatgptimpactgenerative}.  Given the substantial incentives for adversaries to circumvent security measures and obtain responses to otherwise restricted queries, often referred to as ``jailbreak'' attacks, research on security alignment has gained momentum.
Early efforts focused on training-time alignment \cite{glaese2022improvingalignmentdialogueagents, ouyang2022traininglanguagemodelsfollow}, where harmful prompts were introduced during training to adjust the model's behavior to refuse inappropriate requests. Another method involved aligning the model through system prompts, explicitly instructing it to reject harmful commands \cite{bai2022constitutionalaiharmlessnessai}.
While inducing no computational overhead during models' inference, multiple studies have demonstrated that these approaches are insufficient on their own, as simple prompt engineering techniques can effectively circumvent them \cite{wei2023jailbrokendoesllmsafety, qiu2023latentjailbreakbenchmarkevaluating, liu2024jailbreakingchatgptpromptengineering}.
Furthermore, advanced adversarial techniques, such as suffix-based jailbreak attacks \cite{zou2023universaltransferableadversarialattacks} and automatic LLM-assisted jailbreak prompt generation \cite{mehrotra2024treeattacksjailbreakingblackbox, chao2024jailbreakingblackboxlarge}, continue to expose weaknesses in existing defenses.

To address these threats, LLM security research has evolved to include external defenses across all stages of the generation pipeline \cite{Yao_2024}.
We can categorize these defenses into three categories:
(1) Pre-processing, which focuses on filtering harmful inputs before they are processed by the model \cite{jain2023baselinedefensesadversarialattacks, zeng2024shieldgemmagenerativeaicontent}, but at the cost of additional inference time, causing a delay in user-interface systems such as chat-based LLMs.
(2) In-process defenses, which monitor and regulate activations and internal representations during inference \cite{xu2024safedecodingdefendingjailbreakattacks, zhang2025jbshielddefendinglargelanguage, dong2025featureawaremaliciousoutputdetection}. This approach has relatively low computational overhead but is based on a limited number of identified features from the activation space, mostly binary ones, which makes them less robust to a wide range of attacks.
(3) Post-processing, which filters and modifies outputs after generation \cite{phute2024llmselfdefenseself, zeng2024autodefensemultiagentllmdefense} can identify not only harmful inputs but also misaligned models' output; however, it requires processing long segments of text and delays the LLMs' responses.
In addition, different defense methods from all groups are built on additional models, mostly LLMs. These defenses not only increase inference time but also the compute requirement from a system that needs to execute an LLM such as Llamaguard \citep{meta2024llamaguard} as its external defenses.

Therefore, defending against sophisticated attacks remains a challenge, mostly in real-time deployed systems.
To solve these challenges, we propose AlignTree, a lightweight and computationally efficient classifier that enhances the alignment of LLMs and assists in distinguishing between harmful and harmless prompts. Relying solely on base model activations, AlignTree achieves state-of-the-art (SOTA) performance in Attack success rate (ASR) and efficiency, without increasing the refusal rate.
To this end, we rely on two complementary sources of signal: (i) activations projected onto the linear refusal direction following \citet{arditi2024refusallanguagemodelsmediated}, and (ii) motivated by prior work suggesting that refusal behavior in LLMs is not entirely linear  \cite{wollschläger2025geometryrefusallargelanguage, hildebrandt2025refusalbehaviorlargelanguage}, we train non-linear support vector machines (SVMs) with radial basis function (RBF) across tokens and layers' hidden state.

The two types of features are then used to train a Random Forest classifier, which assigns confidence scores reflecting the harmfulness of a prompt. The main advantage of the resulting classifier (AlignTree) is that, in contrast to prior methods, it does not rely on fine-tuning, additional inference passes, or auxiliary models. {Instead, it leverages the LLM’s internal activations to enhance model alignment through targeted probing.} 

We extensively evaluate AlignTree across nine different LLMs and multiple widespread harmfulness benchmarks. AlignTree outperforms existing state-of-the-art defenses by achieving a lower attack success rate (ASR), minimizing unnecessary refusal of harmless instructions, and significantly reducing computational overhead. By addressing the efficiency gaps overlooked in prior work and enabling a more complex defense strategy using confidence scores, AlignTree paves the way for scalable, real-time LLM alignment.

\section{Related work}
Recent advancements in the field of LLMs have significantly enhanced the understanding of their vulnerabilities, defense mechanisms, and security alignment strategies.
\citet{Yao_2024} provided a comprehensive taxonomy of threats and corresponding defenses. LLM inference defenses are often categorized into three stages:  \textbf{Pre-Process}, \textbf{In-Process}, and \textbf{Post-Process}, based on when the defense mechanisms are applied during the model's inference pipeline.

\paragraph{Pre-Process defenses} operate on prompts before they are passed to the LLM for response generation. \citet{jain2023baselinedefensesadversarialattacks} evaluated the effectiveness of different defenses, applying each defense independently to assess its impact.
Perplexity filters, which use the model inference to compute the perplexity score with regard to its input, and potentially output, are designed to identify and filter out gibberish input, such as GCG (Greedy Coordinate Gradient) \cite{zou2023universaltransferableadversarialattacks}. 

The use of LLM-as-a-judge has become a state-of-the-art approach \cite{gu2025surveyllmasajudge}. Security-aligned models, which are usually considered as small LLMs, such as LlamaGuard \cite{meta2024llamaguard} and ShieldGemma \cite{zeng2024shieldgemmagenerativeaicontent}, have proven effective in detecting and assessing harmful inputs.
However, this approach is computationally heavy, requiring storage and execution of an additional LLM and executing additional forward passes.

\paragraph{In-Process defenses}  analyze LLM intermediate results such as neuron activation and hidden states.
\citet{arditi2024refusallanguagemodelsmediated} explored the existence of a refusal direction in hidden states, a single geometric space in the activation space, that can be leveraged to detect and block harmful prompts. Building on this, \citet{zhang2025jbshielddefendinglargelanguage} utilized the refusal direction to identify harmful prompts and then reinforced the awareness of the LLM for the toxic concept via activation addition with the refusal direction. Similarly, \citet{dong2025featureawaremaliciousoutputdetection} trained a binary classifier on refusal direction activations to identify harmful prompts during response generation at every generated token, then steering the model toward producing harmless responses. {Early work primarily treated refusal as a linear phenomenon, using linear directions to fine-tune models or guide their outputs. However, recent research has shown that refusal behavior in LLMs is not entirely linear \citep{hildebrandt2025refusalbehaviorlargelanguage, wollschläger2025geometryrefusallargelanguage}, suggesting that relying solely on linear signals may oversimplify the underlying dynamics and potentially degrade generation quality. In this work, we show that incorporating additional non-linear refusal signals can improve robustness and better mitigate harmful completions.}

Other defenses, such as SmoothLLM \cite{robey2024smoothllmdefendinglargelanguage}, utilize a perturbation technique that copies the prompt and applies small changes to each copy, then generates multiple responses. Using majority voting, the prompt is classified as malicious or not. Similarly, \citet{kumar2025certifyingllmsafetyadversarial} proposed the erase-and-check approach, which involves generating multiple copies of a prompt and randomly removing tokens. The model generates multiple responses, and majority voting is used to determine whether the prompt is malicious. \citet{li2023rainlanguagemodelsalign} proposed {RAIN}, a method that enables models to rewind responses during generation if harmful content is detected. 
These kinds of approaches do not require additional LLM but suffer from a big latency caused by rerunning the base LLM multiple times, especially when considering the fact that in many systems, the ratio of harmful-harmless prompts is low

\paragraph{Post-Process defenses} evaluate the LLM's generated response to harmful content. \citet{phute2024llmselfdefenseself} demonstrated how an LLM can act as a judge to review its responses for potential harm. \citet{zeng2024autodefensemultiagentllmdefense} built on this idea by employing a team of LLM agents that work together through dialogue to evaluate whether a prompt is harmful. \citet{chen2023jailbreakerjailmovingtarget} implemented a multi-metric evaluation system where several LLM judges calculate toxicity and quality metrics before reaching a consensus via majority voting.
These approaches are as strong as the LLM they utilize for the classification of prompts' harmfulness, and require additional compute to host and run. In particular, systems that want to use multi-judges based methods, such as \citet{chen2023jailbreakerjailmovingtarget, zeng2024autodefensemultiagentllmdefense}, dramatically increase the computational requirement for deployed systems.

Table~\ref{Summary overhead} provides an overview of several well-known defense methods and their associated overheads. Unlike other approaches, our method achieves state-of-the-art ASR results without introducing additional inference steps or requiring auxiliary models. In contrast, LlamaGuard and AutoDefense necessitate deploying extra models, leading to increased computational overhead. SmoothLLM and AutoDefense also depend on a large number of prompt variations, which is impractical in real-world scenarios. Self-Defense doubles the inference cost yet still fails to achieve low ASR in most cases. While PerplexityDefense is highly efficient, its simplicity limits its effectiveness against more sophisticated attacks.

\section{Method}
\label{section:Method}
{In this section, we introduce AlignTree, an efficient classifier for detecting harmful responses. AlignTree relies on two complementary signals: (i) scalar features derived from projecting activations onto the model's refusal direction, and (ii) non-linear features extracted by SVMs trained to identify malicious patterns in LLM activations. These signals are then combined and fed into a Random Forest classifier for the final prediction.}

\subsection{Obtaining Refusal Activations}
\label{sec:refusal_activations}
Following \citet{arditi2024refusallanguagemodelsmediated}, we extract a single linear refusal direction $r^*$ that captures the model’s internal representation of refusal. After determining $r^*$, we project hidden states onto this vector to obtain scalar Refusal Activations, which serve as one of the inputs to our classifier.

\subsubsection{Difference-in-means.}
{To detect the single refusal direction}, we begin by constructing a set of \textbf{candidate} refusal directions using the difference-in-means method. Let $D_{\text{harmful}}$ and $D_{\text{harmless}}$ be the sets of harmful and harmless prompts, respectively. For each prompt $t$ in these sets, we extract the hidden activation $x_i^{(l)}(t)$ at token position $i \in \mathcal{I}$ and layer $l \in [L]$ of the LLM, where $L$ is the total number of layers.
We then compute the average activation vectors for each token position and layer over the training subsets $D_{\text{harmful}}^{\text{(train)}}$ and $D_{\text{harmless}}^{\text{(train)}}$:
\begin{align}
  \mu_i^{(l)} &= \frac{1}{|D_{\text{harmful}}^{\text{(train)}}|} \sum_{t \in D_{\text{harmful}}^{\text{(train)}}} x_i^{(l)}(t), \\
  v_i^{(l)} &= \frac{1}{|D_{\text{harmless}}^{\text{(train)}}|} \sum_{t \in D_{\text{harmless}}^{\text{(train)}}} x_i^{(l)}(t),
\end{align}
\noindent
where $x_i^{(l)}(t)$ denotes the hidden activation at position $i$ and layer $l$ for prompt $t$. The difference-in-means vectors are then defined as:
\begin{align}
  r_i^{(l)} = \mu_i^{(l)} - v_i^{(l)}.
\end{align}
This yields a set of candidate directions $\{r_i^{(l)}\}$ across layers and token positions. 

\subsubsection{Selecting a single vector.} We evaluate each candidate vector on held-out validation sets $D_{\text{harmful}}^{\text{(val)}}$ and $D_{\text{harmless}}^{\text{(val)}}$, following the procedure of \citet{arditi2024refusallanguagemodelsmediated}. Each vector is assessed based on its ability to reduce refusal behavior when ablated, and to induce refusal behavior when added, while otherwise preserving the model’s general functionality. The vector with the greatest effect under these criteria is selected as the {single} refusal direction, denoted $r^*$.

\subsubsection{Refusal Activations.}
To measure the alignment of a hidden state $h \in \mathbb{R}^{d_{\text{model}}}$ with the refusal direction, we compute its projection onto $r^*$:
\begin{equation}
\text{proj}_{r^*}(h) = \frac{h \cdot r^*}{\|r^*\|} \in \mathbb{R}
\end{equation}
This scalar value, referred to as the Refusal Activation, measures the degree to which the hidden state aligns with the direction associated with refusal behavior.  
We collect activations from the final token position across multiple layers, resulting in a set of scalar features that together constitute the Refusal Activations.

\subsection{Extracting Non-linear malicious signals}

While a single linear refusal direction captures some aspects of harmful prompt detection, prior work \citep{hildebrandt2025refusalbehaviorlargelanguage, wollschläger2025geometryrefusallargelanguage} suggests that the geometry of refusal in LLMs may be inherently non-linear. To capture richer indicators of harmfulness, we train a large set of Support Vector Machines (SVMs) with radial basis function (RBF) kernels.

For each layer of the model, \(l \in [L]\), and each token position \(i\) among the first 3 and last 5 tokens of the prompt, we train a separate SVM classifier \(\text{SVM}_i^{(l)}\). Each classifier \(\text{SVM}_i^{(l)}\) is trained to distinguish between harmful and harmless prompts using the hidden activations \(x_i^{(l)}(t) \in \mathbb{R}^{d_{\text{model}}}\), taken from a labeled training set.

In total, we train \(8 \times L\) SVMs, one for each combination of the 8 selected token positions and all \(L\) layers. We used the same training set for both model training and Refusal Activation extraction. 
After training, we evaluate all \(8L\) SVMs on a held-out validation set based on accuracy. We then select the top-performing $L/2$ SVMs to use in our classifier.

\subsubsection{Probabilistic Feature Extraction.}
\label{par: probabilisticfeatureextraction}
For each SVM, we use 5-fold cross-validation on the designated training set to generate out-of-fold harmfulness probabilities. To obtain probabilities from the raw SVM, we follow the algorithm by \citet{platt2000probabilistic}, which fits a sigmoid to map decision values to probabilities. This results in a single confidence score per training example for each SVM, enabling us to represent its non-linear signal as a normalized scalar feature used by the final classifier. We denote this calibrated output as \( P_{\text{harmful}}(x_i^{(l)}) \), representing the harmfulness probability predicted by the $\text{SVM}_l^{(l)}$ associated with feature \( i \) at layer \( l \).

Let \(\mathcal{S}\) denote the set of $L/2$ selected classifiers. For a new prompt \(t\), we compute the calibrated harmfulness probabilities of each SVM in \(\mathcal{S}\), resulting in a feature vector of confidence scores that encodes non-linear harmfulness signals:
\begin{align}
    \text{SVMFeatures}(t) = \left[ P_{\text{harmful}}(x_i^{(l)}(t)) \right]_{(i,l) \in \mathcal{S}}.
\end{align}

\subsection{AlignTree}
\label{section:train_AlignTree}
We train a Random Forest classifier using two types of input signals for each prompt \( t \):  
(i) Refusal activations, computed by projecting the final token activations from each layer \( l \in [1..L] \) onto the selected refusal direction \( r^* \); and  
(ii) Harmfulness probability estimates, generated by a selected set \( \mathcal{S} \) of nonlinear SVM classifiers.

The complete input feature vector $F$ is constructed by concatenating these components:
\begin{align}
    \text{F}(t) &= \left[ \text{proj}_{r^*}(x_{-1}^{(l)}(t)) \right]_{l=1}^L 
     \oplus \left[ P_{\text{harmful}}(x_i^{(l)}(t)) \right]_{(i,l) \in \mathcal{S}}
\end{align}
where \( x_{-1}^{(l)}(t) \) denotes the activation at the final token position in layer \( l \), \( x_i^{(l)}(t) \) is the activation at token position \( i \) in layer \( l \), and \( \oplus \) denotes vector concatenation.

To ensure computational efficiency, we employ a lightweight Random Forest model consisting of a small number of shallow decision trees, trained on a curated dataset.

\paragraph{Threshold selection}
\label{par:thresholds}

\begin{figure}[h!]
\centering
\includegraphics[width=0.9\linewidth, clip, trim=0mm 2mm 0mm 0mm]{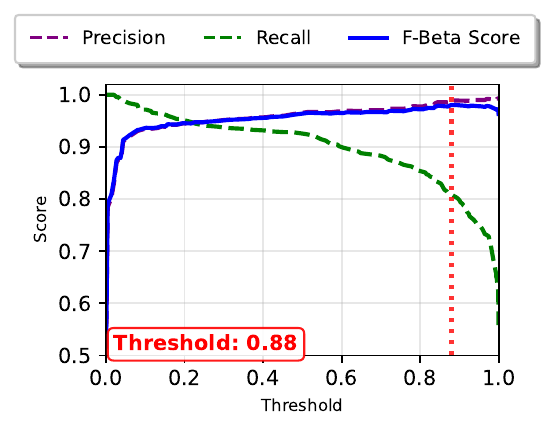}
\caption{Qwen2.5-7B-Instruct threshold selection based on the generalized $F_{\beta}$ score. $\tau = 0.88$. }
\label{fig: threshold_qwen7b}
\end{figure}

{We define a harmfulness threshold \(\tau\) to decide whether a prompt is accepted or blocked. Prompts with predicted harmfulness below \(\tau\) are passed to the LLM, while those above are rejected as malicious. To avoid excessive refusals while minimizing missed harmful prompts, \(\tau\) is selected to maximize precision while balancing recall. This trade-off is optimized using the following \(F_{\beta}\) score:
}
\begin{equation}
    F_\beta = (1 + \beta^2) \cdot \frac{\text{Precision} \cdot \text{Recall}}{(\beta^2 \cdot \text{Precision}) + \text{Recall}}
\end{equation}
To emphasize precision, we set \(\beta = 0.2\). For each model, we select the final threshold as the one that maximizes the \( F_\beta \) score on the validation set.
{See Figure~\ref{fig: threshold_qwen7b} for the generalized F-score curves and the selected threshold for Qwen2.5-7B-Instruct.}
Additional experiments validating the threshold selection are provided in the Appendix.

\begin{table*}[t]
\centering
\begin{small}
\begin{tabular}{llcccc|cccc}
\toprule
Model & Strategy  & \multicolumn{4}{c}{ASR $\downarrow$} & \multicolumn{4}{c}{Refusal $\downarrow$}\\ 
\cmidrule(lr){3-6} \cmidrule(lr){7-10}
& & MalwareGen &PromptInject  & PAIR & AutoDAN & PIQA & OpenBookQA & SIQA & ARC \\ 

\midrule
Qwen2.5 & Baseline & 91.0 & 50.0 & 51.0 & 48.0 & \textbf{0} & \textbf{0} & \textbf{0} & \textbf{0} \\ 
-0.5B &  AutoDefense & 5.0 & \textbf{0} & 13.0 & \textbf{0} & \textbf{0} & 6.0 & 3.0 & 8.0 \\ 
-Instruct & SelfDefense-Input & 43.0 & 13.0 & 8.0 & 17.0 & 80.0 & 35.0 & 33.0 & 46.0 \\ 
 & SelfDefense & 42.0 & 16.0 & 11.0 & 13.0 & 72.0 & 37.0 & 37.0 & 41.0 \\ 
 & PerplexityDefense & 84.0 & 50.0 & 50.0 & 47.0 & \textbf{0} & \textbf{0} & \textbf{0} & \textbf{0} \\ 
 & SmoothLLM & 77.0 & 43.0 & 49.0 & 44.0 & \textbf{0} & \textbf{0} & \textbf{0} & \textbf{0} \\ 
 & AlignTree (\textbf{Ours}) & \textbf{4.0} & 41.0 & \textbf{6.0} & \textbf{0} & \textbf{0} & \textbf{0} & \textbf{0} & \textbf{0} \\ 
\midrule 
Llama & Baseline & 9.0 & 43.0 & 14.0 & \textbf{0} & 2.0 & \textbf{0} & \textbf{5.0} & \textbf{0} \\ 
-3.1-8B &  AutoDefense & \textbf{5.0} & \textbf{0} & 16.0 & \textbf{0} & 2.0 & 1.0 & 7.0 & 4.0 \\ 
-Instruct & SelfDefense-Input & 8.0 & 32.0 & \textbf{8.0} & \textbf{0} & 55.0 & 47.0 & 30.0 & 52.0 \\ 
 & SelfDefense & 8.0 & 28.0 & \textbf{8.0} & \textbf{0} & 51.0 & 55.0 & 34.0 & 49.0 \\ 
 & PerplexityDefense & 8.0 & 42.0 & 15.0 & \textbf{0} & 2.0 & \textbf{0} & \textbf{5.0} & \textbf{0} \\ 
 & SmoothLLM & 8.0 & 37.0 & 13.0 & \textbf{0} & 2.0 & \textbf{0} & \textbf{5.0} & \textbf{0} \\ 
 & AlignTree (\textbf{Ours}) & \textbf{5.0} & 18.0 & 9.0 & \textbf{0} & \textbf{1.0} & \textbf{0} & \textbf{5.0} & \textbf{0} \\ 
\midrule 
gemma & Baseline & 24.0 & 50.0 & 36.0 & 6.0 & \textbf{0} & \textbf{0} & \textbf{0} & \textbf{0} \\ 
-3-12b &  AutoDefense & \textbf{7.0} & \textbf{5.0} & 19.0 & \textbf{1.0} & \textbf{0} & 7.0 & \textbf{0} & 2.0 \\ 
-it & SelfDefense-Input & 23.0 & 35.0 & 28.0 & 3.0 & 1.0 & \textbf{0} & 1.0 & \textbf{0} \\ 
 & SelfDefense & 18.0 & \textbf{5.0} & 33.0 & 4.0 & 2.0 & 17.0 & 58.0 & 11.0 \\ 
 & PerplexityDefense & 16.0 & 52.0 & 35.0 & 5.0 & \textbf{0} & \textbf{0} & \textbf{0} & \textbf{0} \\ 
 & SmoothLLM & 25.0 & 55.0 & 37.0 & 7.0 & \textbf{0} & \textbf{0} & \textbf{0} & \textbf{0} \\ 
 & AlignTree (\textbf{Ours}) & 10.0 & 40.0 & \textbf{10.0} & \textbf{1.0} & \textbf{0} & \textbf{0} & \textbf{0} & \textbf{0} \\
\bottomrule
\end{tabular}
\end{small}
\caption{Attack Success Rate (ASR) for each harmful dataset and model, as well as Refusal rates for harmless datasets. The full results (nine LLMs from three families) are in the Appendix and show similar patterns. }
\label{tab:main results}
\end{table*}
\section{Experiments}
\label{sec:experiments} 
\noindent\textbf{Refusal and SVM Datasets.} In our experiments, we compile two datasets for training the refusal vectors and SVMs: (i) $D_{harmful}$: Prompts labeled as harmful, drawn from {Advbench} \cite{zou2023universaltransferableadversarialattacks}, {MaliciousInstruct} \cite{huang2023catastrophicjailbreakopensourcellms}, {TDC2023} \cite{mantas2023tdc}, {StrongReject} \cite{souly2024strongrejectjailbreaks} and {HarmBench} \cite{mazeika2024harmbenchstandardizedevaluationframework}. (ii) $D_{harmless}$: a collection of benign prompts sampled from {ALPACA} \cite{alpaca}. Additionally, we included the white-box targeted attack from \citet{adaptivejailbreaking2025} to further evaluate AlignTree's robustness, with results detailed in the Appendix.

\par
\noindent\textbf{Random Forest Datasets.} In addition, we construct an additional dataset for training the Random Forest classifier, combining harmful datasets: Jailbreakbench \cite{patrick2024jailbreakbench}, PAIR \cite{chao2024jailbreakingblackboxlarge}, and AutoDAN \cite{liu2024autodangeneratingstealthyjailbreak} attack samples sourced from \citet{zhang2025jbshielddefendinglargelanguage}, along with the harmless datasets ALPACA \citet{alpaca} and {XSTest} \cite{rottger-etal-2024-xstest}.  Samples from ALPACA are present in both our Random Forest training set and the Refusal and SVM datasets; however, we made sure that there is no overlap between these datasets. 
\par
\noindent\textbf{AlignTree Hyperparameters.} The Random Forest classifier uses the hyperparameters $n\_estimators=50$, $max\_depth=6$, and $min\_samples\_split=5$. A grid search exploring hyperparameter sensitivity is detailed in the Appendix. The SVMs employ an RBF kernel; the use of linear kernels is explored in our Ablation Study. Feature importance for each signal is detailed in the Appendix.
\par
\noindent\textbf{Evaluation datasets.} For evaluating adversarial robustness, we primarily leveraged {\bf harmful datasets} from {Garak} \cite{derczynski2024garakframeworksecurityprobing}, including:  
{PromptInject}, an implementation of \citet{ignore_previous_prompt}, designed to disrupt model alignment using directives such as \textit{``IGNORE PREVIOUS INSTRUCTIONS''}, and {MalwareGen}, a collection of prompts aimed at eliciting malware-generating code from the model.  Additionally, we incorporated two attack datasets from {JBShield} \cite{zhang2025jbshielddefendinglargelanguage} that bypassed the safeguards of {Llama-3.1-8B}:  {PAIR}, an algorithm that crafts semantic jailbreaks using only black-box access to an LLM \cite{chao2024jailbreakingblackboxlarge}, and AutoDAN, a dataset of adversarial attacks generated via genetic algorithms, requiring only black-box access to an LLM \cite{liu2024autodangeneratingstealthyjailbreak}.  {Samples from PAIR and AutoDAN are included in both our Random Forest training dataset and evaluation datasets; however, we ensured that there is no overlap between them.}

To ensure that AlignTree does not degrade performance or lead to excessive refusals of {\bf harmless} responses, we evaluated it on four benign, commonsense reasoning datasets:  

    {PIQA} \cite{Bisk2020} -– assessing physical commonsense reasoning;  
    {ARC-Challenge} \cite{clark2018thinksolvedquestionanswering} –- testing scientific reasoning;
     {OpenBookQA} \cite{OpenBookQA2018} –- evaluating advanced question answering; and   
    {SIQA} (Social Interaction QA) \cite{siqa2019} -– measuring social commonsense understanding.  

\begin{figure*}[h!]
\centering
\centerline{\includegraphics[width=1\linewidth]{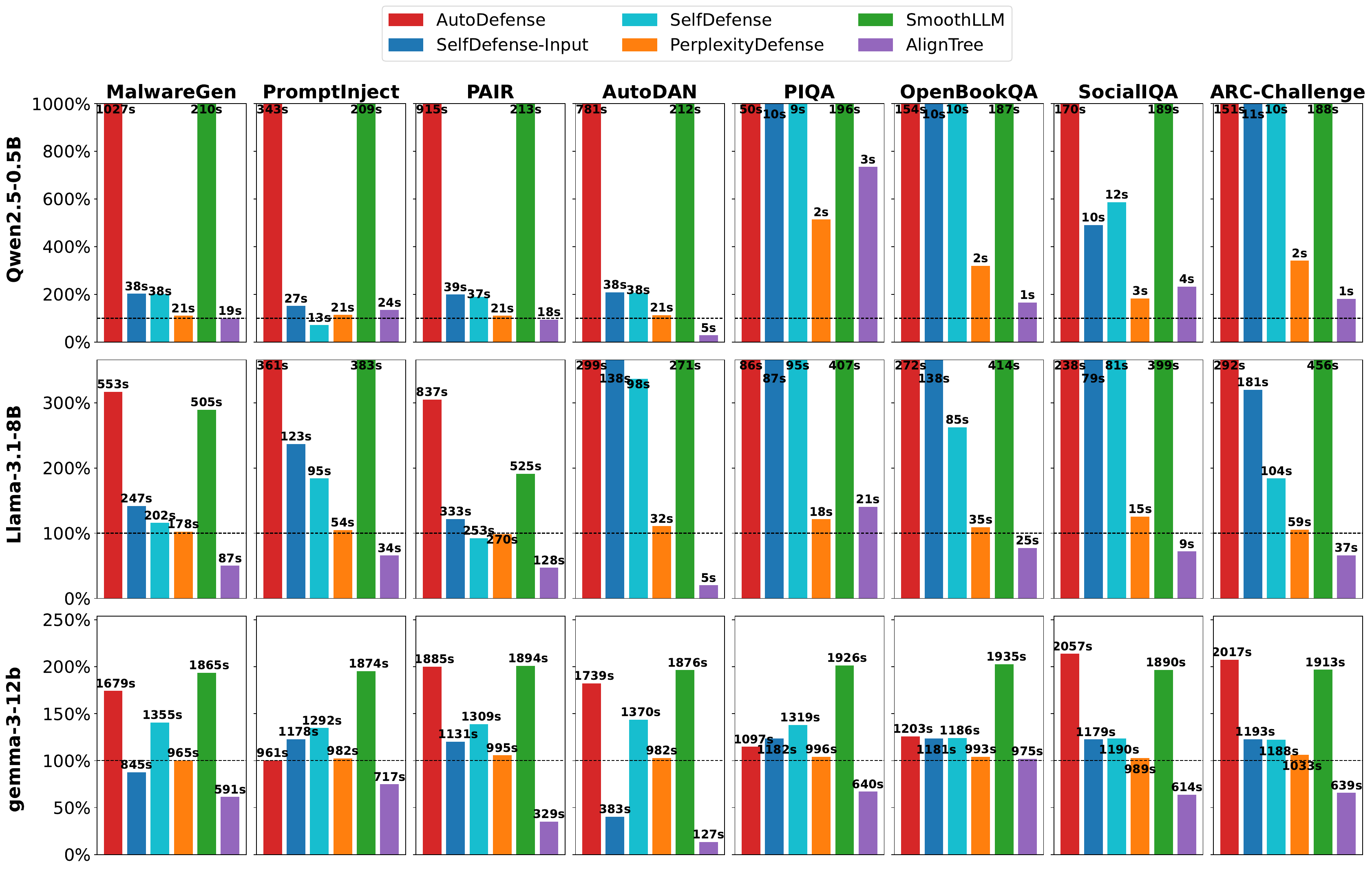}}
\caption{Execution time per method relative to running the baseline LM (dashed line) (Lower is better). Charts are capped at 1000\% of baseline time. The full results (nine LLMs from three families) are in the Appendix and show similar patterns.} 

\label{fig: barchart}
\vspace{6px}
\end{figure*}

\noindent{\bf Baselines\quad} To assess the effectiveness of our efficient classifier, we compare it against eight defense strategies, including several state-of-the-art methods. We focus on methods that do not require deploying auxiliary models. These strategies include: (i) {Baseline} which relies solely on the model's native alignment; (ii) {AutoDefense} \cite{zeng2024autodefensemultiagentllmdefense},  (iii) {SmoothLLM} \cite{robey2024smoothllmdefendinglargelanguage}, (iv) {SelfDefenseInput} and (v) {SelfDefense} \cite{phute2024llmselfdefenseself}, which query the main model on the harmfulness of the prompt and response, respectively; (vi) PerplexityDefense \cite{jain2023baselinedefensesadversarialattacks}.
For these defenses, we chose the hyperparameters per their original papers and are described in the Appendix, as well as additional implementation details.

We evaluate three families of instruction-tuned LLMs: {Qwen2.5} (0.5B, 3B, 7B) \cite{qwen2.5}, {Llama3} (1B, 3B, 8B) \cite{grattafiori2024llama3herdmodels}, and {Gemma3} (1B, 4B, 12B) \cite{gemmateam2025gemma3technicalreport}. 
In this section, we present results for a single model per family to maintain clarity, selecting different sizes to ensure diversity: Qwen2.5-0.5B-Instruct, Llama-3.1-8B-Instruct, and Gemma-3-12B-It. Complete results for all nine models are included in the Appendix and exhibit the same patterns as those presented in the main text.

{Following prior work \cite{mazeika2024harmbenchstandardizedevaluationframework, arditi2024refusallanguagemodelsmediated, zhang2025jbshielddefendinglargelanguage}, we adopt the Attack Success Rate (ASR) metric, which measures the proportion of harmful completions that bypass refusal mechanisms. To evaluate both harmfulness and refusals, we rely on ChatGPT-4o \cite{openai2024chatgpt4o}, using its responses and a set of refusal-related keywords.
In addition to adversarial evaluation, we conduct a complementary experiment on benign datasets to measure over-refusal and execution time.

First, we assess the trade-off between \textbf{attack success rate (ASR) and refusal behavior} using both harmful and harmless prompt datasets. We evaluate each defense’s ability to block harmful prompts while minimizing refusals of harmless ones, ensuring practical real-world applicability. Table~\ref{tab:main results} presents ASR and refusal results for representative model families and sizes; full results in the Appendix exhibit the same trends. 
AlignTree demonstrates robust performance across all evaluated models and datasets, achieving substantial reductions in ASR compared to the no-defense baseline while maintaining the lowest refusal rates.
In all tested scenarios, it delivers state-of-the-art refusal performance, showing the lowest rates across datasets and model families.

Across most datasets, AlignTree matches or exceeds the ASR performance of existing defenses, including more complex approaches such as AutoDefense and SelfDefense. For instance, on Gemma-3-12B, it attains the lowest ASR for the PAIR dataset; on Qwen2.5-0.5B, it records the lowest ASR among MalwareGen, PAIR, and AutoDAN.
It also performs competitively on Llama-3.1-8B, closely matching or surpassing other defenses across all datasets. However, there are cases where AlignTree shows higher ASR than other defenses, for example, on PromptInject with Qwen2.5-0.5B and Gemma-3-12B, SelfDefense and AutoDefense achieve lower ASR at the cost of higher refusal rates, which frequently block benign inputs and cause over-refusal behavior. In contrast, AlignTree provides strong protection while minimizing unnecessary refusals.

Secondly, we evaluate \textbf{defenses' efficiency}, defining execution time as the total duration required to process 100 prompts from a given task.
Figure~\ref{fig: barchart} shows that for most models and datasets, AlignTree achieves the lowest execution time. The only exceptions are a few cases where PerplexityDefense is marginally faster; however, PerplexityDefense incurs a higher ASR.
Notably, AlignTree’s execution time remains highly competitive with the baseline methods, introducing only negligible overhead compared to other defenses.

{In summary, AlignTree delivers the strongest overall performance by combining substantial ASR reductions with the lowest refusal rates, while also achieving state-of-the-art execution times across most models and datasets. This balance of robustness, low refusal, and computational efficiency makes AlignTree a dependable and practical defense across diverse models and threat environments.

\begin{table*}[h!]
\centering
\begin{small}
\begin{tabular}{@{}lllcccccccc@{}}
\toprule
 Model  & Strategy & \multicolumn{2}{c}{MalwareGen} & \multicolumn{2}{c}{PromptInject} & \multicolumn{2}{c}{PIQA} 
 &  \multicolumn{2}{c}{ARC-Challenge}\\\cmidrule(lr){3-4} \cmidrule(lr){5-6} \cmidrule(lr){7-8} \cmidrule(lr){9-10}
 & & ASR $\downarrow$ & Time $\downarrow$ & ASR $\downarrow$ & Time $\downarrow$ & Refusal $\downarrow$ & Time $\downarrow$ & Refusal $\downarrow$ & Time $\downarrow$ \\
\midrule
Qwen2.5 & RefusalClassifier & 89.0 & 27.18s & 52.0 & 27.34s & \textbf{0} & 1.44s & \textbf{0} & 1.79s \\ 
-0.5B & SVMClassifier & 33.0 & 21.85s & 46.0 & 27.05s & \textbf{0} & 1.46s & \textbf{0} & 1.46s \\ 
-Instruct & MultiRefusalsClassifier & 29.0 & \textbf{17.45}s & 53.0 & 18.32s & \textbf{0} & \textbf{0.58}s & \textbf{0} & \textbf{0.89}s \\ 
 & AlignTreeLinear & 61.0 & 22.67s & 43.0 & \textbf{16.57}s & \textbf{0} & 0.95s & \textbf{0} & 1.09s \\ 
 & AlignTree & \textbf{4.0} & 19.01s & \textbf{41.0} & 24.8s & \textbf{0} & 3.16s & \textbf{0} & 1.27s \\ 
\midrule 
Llama & RefusalClassifier & 5.0 & 145.54s & 44.0 & 57.68s & 2.0 & 20.55s & \textbf{0} & 59.84s \\ 
-3.1-8B & SVMClassifier & \textbf{2.0} & 66.58s & 20.0 & 42.02s & \textbf{1.0} & 17.02s & \textbf{0} & 43.9s \\ 
-Instruct & MultiRefusalsClassifier & 4.0 & \textbf{62.49}s & 32.0 & \textbf{31.67}s & \textbf{1.0} & \textbf{11.66}s & \textbf{0} & \textbf{35.79}s \\ 
 & AlignTreeLinear & 7.0 & 101.99s & \textbf{18.0} & 40.61s & \textbf{1.0} & 14.38s & \textbf{0} & 36.74s \\ 
 & AlignTree & 5.0 & 87.37s & \textbf{18.0} & 34.2s & \textbf{1.0} & 21.88s & \textbf{0} & 37.44s \\ 
\midrule 
gemma & RefusalClassifier & 21.0 & 619.41s & 54.0 & 957.3s & \textbf{0} & 981.92s & \textbf{0} & 983.69s \\ 
-3-12b & SVMClassifier & 26.0 & 738.11s & 37.0 & 708.4s & 5.0 & 965.68s & \textbf{0} & 969.49s \\ 
-it & MultiRefusalsClassifier & 25.0 & 509.97s & 53.0 & \textbf{602.4}s & \textbf{0} & \textbf{640.47}s & \textbf{0} & \textbf{606.22}s \\ 
 & AlignTreeLinear & \textbf{8.0} & \textbf{496.35}s & \textbf{29.0} & 800.97s & \textbf{0} & 959.48s & \textbf{0} & 949.66s \\ 
 & AlignTree & 10.0 & 591.11s & 40.0 & 717.22s & \textbf{0} & 988.4s & \textbf{0} & 978.93s \\ 
\bottomrule
\end{tabular}
\end{small}
\caption{This table reports ASR, refusal rates, and execution time for each dataset, illustrating the impact of ablating individual components of AlignTree. The full results across eight datasets and nine LLMs from the three families are in the Appendix.}
\label{tab:ablation_study}
\end{table*}

\subsection{Ablation study}
\label{sec:ablation}
In this experiment, we evaluate the contribution of each signal by independently training separate Random Forest classifiers under four configurations. Our goal is to verify that combining these signals yields superior performance compared to any individual component: (i) RefusalClassifier — trained solely on the activations from a single refusal vector without SVM signals; (ii) SVMClassifier — trained only on non-linear SVM decision boundaries without incorporating refusal activations; (iii) MultiRefusalsClassifier — leveraging activations from multiple top-performing refusal vectors across layers and tokens; and (iv) AlignTreeLinear — using a single refusal vector with SVMs constrained to linear decision boundaries.

The complete AlignTree method delivers the most consistent performance across all evaluated models and datasets, striking a strong balance between low ASR and efficient execution time, without increasing refusal rates.
For instance, on Qwen2.5-0.5B, AlignTree achieves the lowest ASR across all datasets while maintaining a competitive runtime. On Llama-3.1-8B, it attains the lowest ASR on PromptInject and closely matches the top results on MalwareGen. The primary exception is Gemma-3-12b, where AlignTreeLinear outperforms both AlignTree and all other defenses in terms of ASR.
Despite this isolated advantage, AlignTreeLinear exhibits significantly worse performance in other settings, such as an ASR of 61.0 on MalwareGen for Qwen2.5-0.5B, compared to just 4.0 ASR with AlignTree.
While it benefits from slightly faster execution, its reliance on linear classifiers limits expressiveness and leads to inconsistent results. 
The SVMClassifier, although leveraging non-linear signals, fails to generalize across datasets and exhibits excessive refusal rates, particularly on Gemma-3-12b.

Some variants, such as the RefusalClassifier and the MultiRefusalsClassifier, exhibit substantial ASR variability: for example, they perform strongly on Llama-3.1-8B-Instruct but poorly on Qwen2.5-0.5B-Instruct. We attribute this inconsistency to differences in the base models’ pretrained alignment behavior, which we discuss further in the Appendix. Nevertheless, the MultiRefusalsClassifier outperforms its single-classifier counterpart, reinforcing the hypothesis that refusal mechanisms are multidimensional phenomena. 

Most classifiers manage to avoid over-refusal, preserving usability on benign datasets such as PIQA and ARC. Notable exceptions include the SVMClassifier on PIQA for Gemma-3-12b and the RefusalClassifier on PIQA for Llama-3.1-8B, both of which demonstrate elevated refusal rates.

In summary, AlignTree emerges as the most reliable and general-purpose defense, consistently achieving a favorable trade-off between robustness, efficiency, and usability. While other classifier-based defenses leveraging model activations may be suitable in certain contexts, AlignTree demonstrates the most stable and dependable performance overall, making it a strong candidate for real-world deployment. Additional ablation results across all models and datasets are provided in the Appendix, demonstrating similar trends.

\section{Conclusions}
We introduced AlignTree, an efficient defense that enhances model alignment while maintaining minimal computational overhead.
In order to build this lightweight deference, we trained a Random Forest classifier that integrates the linear refusal direction with a novel SVM-based signal designed to capture non-linear features associated with harmful content. Our results show that AlignTree consistently outperforms existing defenses in terms of ASR, refusal, and computational efficiency while introducing a non-negligible increase in execution time over the baseline. Moreover, our results demonstrate that leveraging non-linear harmfulness signals leads to improved alignment performance compared to relying solely on a single linear refusal vector, which we believe is essential for advancing alignment strategies. In future work, we plan to extend AlignTree by introducing an additional ``suspicious'' threshold, one that distinguishes borderline prompts from clearly benign or harmful ones. It will allow identifying prompts that warrant further analysis without immediate rejection and can be used jointly with additional defenses.

\section*{Limitations}
While AlignTree represents a meaningful advancement in improving the alignment of LLMs, several limitations remain.

ASR evaluations in this work were conducted using another LLM, following methodologies similar to those in  \citet{arditi2024refusallanguagemodelsmediated} and related defense studies. While practical, this approach may occasionally introduce evaluation inaccuracies due to model-based judgment. 

Another limitation is that AlignTree requires training a separate classifier for each model, and its effectiveness depends heavily on the level of the base model's initial alignment and the quality of the data. Finally, while this work combined linear and non-linear signals, further research could explore more direct approaches to characterizing and utilizing non-linear refusal properties; for instance, by identifying additional semantic directions or better modeling the refusal manifold in latent space.

{Finally, AlignTree relies on a limited set of input signals and lightweight classifiers to reduce the risk of overfitting. Future work could explore the use of more complex models and larger training datasets to further enhance performance.}
\smallskip
\section*{Ethics Statement}
This work aims to enhance language models by introducing a novel method to improve their safe usage through efficient and robust defenses. We recognize the potential of such technologies and emphasize the importance of their responsible use. While our contributions are intended to support the development of more aligned models, we stress the need to prevent misuse, such as generating harmful content. Future research should focus on promoting more efficient defense strategies that align with societal benefits.

  \section*{Acknowledgements}
    This work was supported by a Tel Aviv University Center for AI and Data Science (TAD) grant. This research was also supported by the Ministry of Innovation, Science \& Technology, Israel (1001576154) and the Michael J. Fox Foundation (MJFF-022407).
    The contribution of SK is part of a PhD thesis research conducted at Tel Aviv University.
    
\bibliography{custom}
\appendix

\newpage

\section{Additional experiment details}
\label{sec:implementation_details}

\subsection{Random Forest hyperparameter sensitivity}
In this section, we demonstrate that our {Random Forest (RF) classifier} exhibits low sensitivity to hyperparameter variations. We conducted a rigorous {grid search} encompassing ${36}$ distinct hyperparameter configurations. Specifically, the search space was defined by the following sets of values: $\mathbf{n\_estimators} \in \{30, 50, 70\}$, $\mathbf{max\_depth} \in \{4, 6\}$, $\mathbf{min\_samples\_leaf} \in \{2, 5, 10\}$, and $\mathbf{min\_samples\_split} \in \{3, 5\}$.

Evaluation utilized a representative set of models, scaling from {Qwen2.5-0.5B-Instruct} to {Llama-3.1-8B-Instruct}, selecting one from each family. Table~\ref{appx: rf-hp} confirms high stability across hyperparameter settings: {Refusal} rates were near $\mathbf{0}$, and {Execution Time} variance was minimal compared to the mean. While {ASR} showed the most variance, results remain significantly better than other SOTA defenses.
\begin{table}[h!]
\centering
\small
\begin{tabular}{lll}
\toprule
Model & \multicolumn{2}{c}{PAIR} \\\cmidrule(lr){2-3} 
&  ASR $\downarrow$ & Time $\downarrow$  \\

\midrule
Qwen2.5-0.5B-Instruct & 14.14 $\pm$ 5.44 & 18.01 $\pm$ 3.44 \\
gemma-3-4b-it & 14.19 $\pm$ 4.93 & 111.97s $\pm$ 8.17 \\
Llama-3.1-8B-Instruct & 8.36 $\pm$ 3.79 & 99.39 $\pm$ 27.39 \\
\midrule 
\midrule 
Model & \multicolumn{2}{c}{PIQA} \\\cmidrule(lr){2-3} 
&  Refusal $\downarrow$ & Time $\downarrow$ \\
\toprule
Qwen2.5-0.5B-Instruct & 0 $\pm$ 0 & 1.11 $\pm$ 0.08 \\
gemma-3-4b-it & 0 $\pm$ 0 & 112.24s $\pm$ 3.15s\\
Llama-3.1-8B-Instruct & 1.14 $\pm$ 0.59 & 12.09 $\pm$ 0.73\\

\bottomrule
\end{tabular}
\caption{Results of the {Random Forest Hyperparameter Grid Search} on the {PAIR} (Harmful) and {PIQA} (Harmless) evaluation datasets.}
\label{appx: rf-hp}
\end{table}

\subsection{AlignTree test results}
\label{appx:AlignTree_test}
In this subsection, we provide the test results on the Random Forest test set, derived from the dataset outlined in the paper.
As illustrated in Figure~\ref{fig: aligntree_conf}, AlignTree delivers an accuracy of 98.86\%, with precision at 98.96\%. This demonstrates that our $F_\beta$ threshold functions effectively on the test set and avoid overfitting to the validation or training datasets.

\begin{figure}[!ht]
\centering
\centerline{\includegraphics[width=1\linewidth]{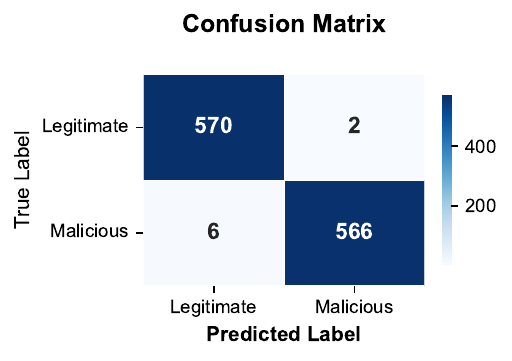}}
\caption{Confusion matrix on the test set for Llama-3.1-8B-Instruct.}
\label{fig: aligntree_conf}
\vspace{8px}
\end{figure}

\subsection{Evaluation Against Adaptive White-Box Attacks}
To comprehensively test AlignTree's robustness against more targeted attacks, we employed an adaptive white-box jailbreak attack methodology based on \citet{adaptivejailbreaking2025}. This attack uses an adversarial prompt template combined with a random suffix search, a targeted technique that subverts safety alignment by observing model activations.

We ran this evaluation on the largest model from each family (Llama-3.1-8B-Instruct, Qwen-2.5-7B-Instruct, gemma-3-12b-it). Using the attack's provided base instructions, we filtered those found in our existing datasets, resulting in 39 unique prompts. For each model and prompt, we executed the full attack process to generate a targeted adversarial prompt capable of jailbreaking the model.

As summarized in Table~\ref{tab:adaptive attack}, AlignTree demonstrates exceptional performance against this targeted attack, achieving $0\%$ ASR on all tested models. While AutoDefense also reaches $0\%$ ASR, it requires a substantially higher computational cost (over $3\times$ the baseline execution time). This confirms AlignTree's superior ability to withstand targeted attacks compared to existing SOTA defenses, which exhibited significantly higher ASRs much closer to the baseline.

\begin{table}[h!]
\centering
\small
\begin{tabular}{llll}
\toprule
model name & strategy & ASR & Time \\
\midrule
Llama-3.1-8B-Instruct & Baseline & 12.82 & 45.02s \\
 & SelfDefenseInput & 5.13 & 37.23s \\
 & SelfDefense & 7.69 & 65.86s \\
 & SmoothLLM & 12.82 & 207.36s \\
& PerplexityDefense & 12.82 & 43.59s \\
& AutoDefense & \textbf{0} & 140.74s \\
 & AlignTree & \textbf{0} & \textbf{2.40}s \\
\midrule
Qwen2.5-7B-Instruct & Baseline & 10.26 & 66.93s \\
 & SelfDefenseInput & 10.26 & 81.99s \\
 & SelfDefense & 10.26 & 81.65s \\
 & SmoothLLM & 10.26 & 142.11s \\
 & PerplexityDefense & 10.26 & 66.76s \\
 & AutoDefense & \textbf{0} & 916.62s \\
 & AlignTree & \textbf{0} &\textbf{ 1.04}s \\
\midrule
gemma-3-12b-it & Baseline & 10.26 & 256.54s \\
 & SelfDefenseInput & \textbf{0} & 60.01s \\
 & SelfDefense & 7.69 & 319.68s \\
 & SmoothLLM & 10.26 & 469.50s \\
 & PerplexityDefense & 10.26 & 270.12s \\
 & AutoDefense & \textbf{0} & 830.28s \\
 & AlignTree & \textbf{0} &\textbf{ 3.12}s \\
\bottomrule
\end{tabular}
\caption{ASR and Time results on the white box Adaptive attack with 39 prompts.}
\label{tab:adaptive attack}
\end{table}

\subsection{Defense hyperparameters}
\label{app:defense_hyperparameters}
SmoothLLM \cite{robey2024smoothllmdefendinglargelanguage}  hyperprameters - We utilized the {RandomSwapPerturbation} setting with parameters {num\_copies}$=10$ and {pert\_pct}$=10$ as recommended by the original paper. This entails generating 10 copies of the prompt, and at each copy, swap $10\%$ of the tokens with tokens selected uniformly from all printable strings then generate a response for each copy and evaluate if the LLM refused to answer, if majority of copies are refusing to answer then this defense considers the prompt to be harmful.

AutoDefense \cite{zeng2024autodefensemultiagentllmdefense} Hyperparameters – We adopted the same hyperparameters as those used in the original paper, implementing a 3-agent setup consisting of a ``Judge'' agent, an ``Intention Analyzer'' agent, and a ``Prompt Analyzer'' agent. Additionally, we used the prompts provided in the official code base, with one minor modification: we instructed the agents not to return a harmful verdict when there is insufficient context. This adjustment addressed a recurring issue we observed, where the system frequently refused outputs unnecessarily.

\begin{figure*}[h!]
\centering
\centerline{\includegraphics[width=1\linewidth]{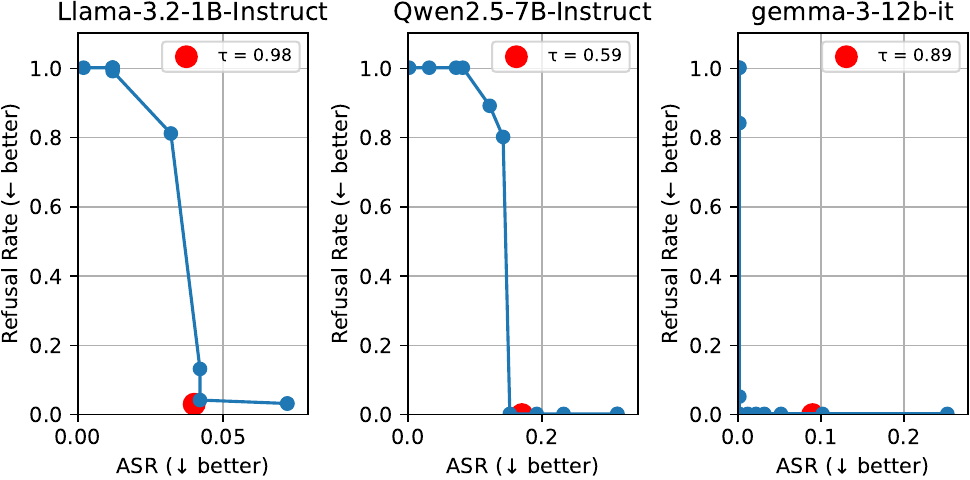}}
\caption{AlignTree Hyperparameters. Threshold performance distributions based on ASR and Refusal metrics for Llama-3.2-1B-Instruct, Qwen2.5-7B-Instruct, and Gemma3-12B-it. Each value in the chart corresponds to a threshold sampled at 0.1 intervals from 0 to 1. The threshold selected $\tau$ using the $F_\beta$-score for balancing precision and recall is highlighted in red.}
\label{fig: hyperparameters}
\end{figure*}

\subsection{Threshold}
\subsubsection{Threshold selection}
We previously stated that thresholds are selected based on the generalized $F_\beta$ score, defined as:

\begin{equation}
F_\beta^{(\text{precision})} = (1 + \beta^2) \cdot \frac{\text{Precision} \cdot \text{Recall}}{\beta^2 \cdot \text{Precision} + \text{Recall}}
\end{equation}

In this section, we report the resulting thresholds for all models. Figure~\ref{fig: all_model_thresholds} displays the $F_\beta$ curves alongside the selected thresholds for each of the nine evaluated models.

\begin{figure*}[h!]
\centering
\centerline{\includegraphics[width=1\linewidth]{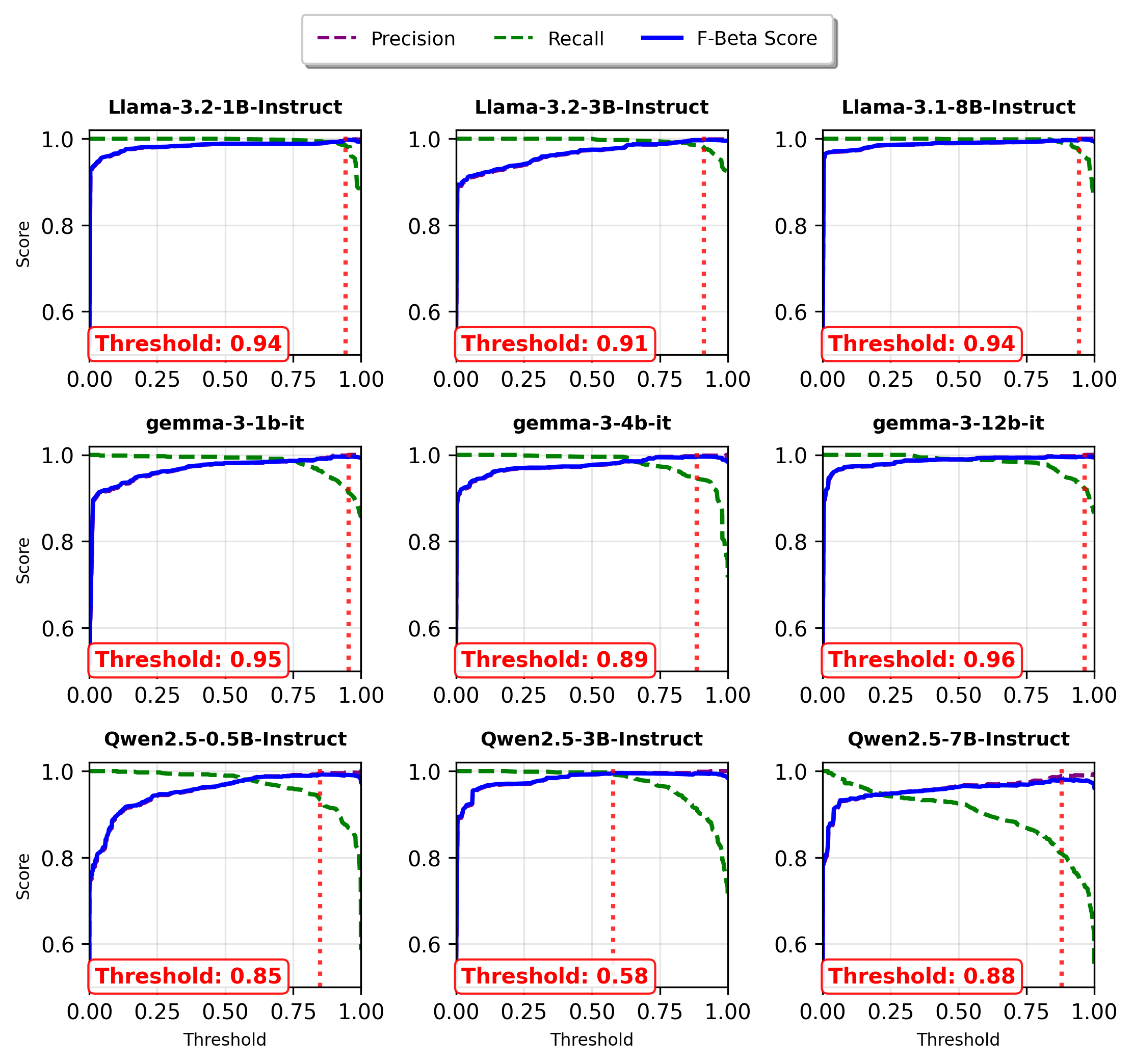}}
\caption{Selected thresholds for each of the models based on the $F_\beta$ score that prioritizes precision but balances with recall.}
\label{fig: all_model_thresholds}
\end{figure*}

\subsubsection{Threshold sensitivity}
\label{app:threshold_hyperparameters}
In this section, we also examined different values for, $\tau$ to verify our method and to examine sensitivity.

Figure~\ref{fig: hyperparameters} shows the results. While the generalized $F_\beta$ score was used to guide hyperparameter selection, the chosen configurations were consistently near-optimal. For Llama-3,2-1B-Instruct, the selected threshold outperformed all others. In Qwen2.5-7B-Instruct, the chosen threshold of 0.59 closely approximated the optimal value of 0.6. These results indicate that the precision-recall based selection effectively meets our objectives-- successfully filtering malicious prompts without inducing excessive refusals.

\subsection{Refusal and Jailbreak evaluations}
\label{appendix_evaluation}
We build on previous work to efficiently analyze LLM refusals, leveraging common ``refusal substrings'' found across models. The specific substrings employed are listed in Table~\ref{tab:refusal_expressions}. In addition to refusal detection, we use ChatGPT-4o to assess the harmfulness of the response and determine whether the malicious request was successfully answered.

\begin{table*}[h!]
    \centering
    \begin{tabular}{|l|}
    \hline
    \rowcolor[gray]{0.9} "I'm sorry", "I am sorry", "I apologize", "As an AI", "As a language model", \\
    \rowcolor[gray]{0.9} "As an assistant", "I cannot", "I can't", "I am unable to", "I'm unable to", \\
    \rowcolor[gray]{0.9}"I am not able to",  "I'm not able to" \\
    \hline
    \end{tabular}
    \caption{Refusal Expressions}
    \label{tab:refusal_expressions}
\end{table*}

\subsection{Category-specific Breakdown of ASR}
To provide an alternative view of AlignTree's robustness, we further break down the Attack Success Rate (ASR) by categories adopted from  \citet{patrick2024jailbreakbench}. We classified each harmful instruction in our datasets into the following categories: Physical Harm (Phys harm), Malware/Hacking (Hack), Economical harm (Economic), Disinformation (Disinfo), Harassment/Discrimination (Harass/Discri), Privacy (Priv), and Fraud/Deception (Decept).

The comparison for the Baseline and AlignTree defense strategies is presented in Figure~\ref{fig: domain-radar}. The results show that AlignTree is able to consistently improve upon the baseline ASR across all these harmful categories. Crucially, there are no significantly worse categories where AlignTree performs poorly, indicating that our method provides consistent robustness across diverse types of harmful content.

\subsection{AlignTree Feature importance}
In this section, we present the top 10 most important features for Llama-3.2-1B-Instruct, Qwen2.5-7B-Instruct, and Gemma-3-12B-IT. The leading features are shown in Figure~\ref{fig: feature_importance}. Notably, SVM-based signals are consistently utilized, and the most influential layers tend to be located in the middle of the model.

\begin{figure*}[b]
\centering
\centerline{\includegraphics[width=1\linewidth]{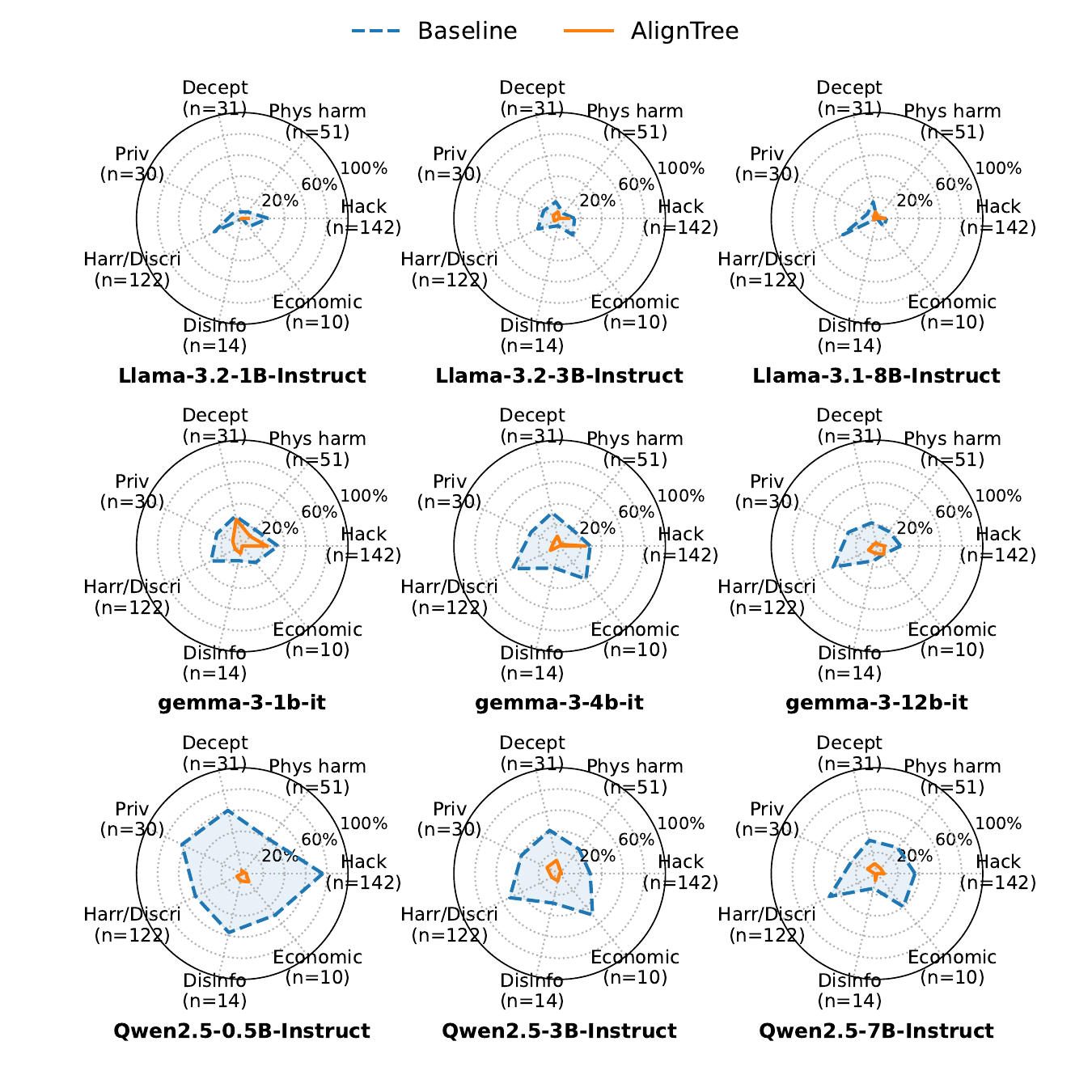}}
\caption{Comparison of Attack Success Rate (ASR) for Baseline and AlignTree across harmful domains, where a lower ASR indicates higher robustness; the number of examples for each harmful domain is shown as $\mathbf{(n=x)}$.}
\label{fig: domain-radar}

\end{figure*}

\begin{figure*}[b]
\centering
\centerline{\includegraphics[width=1\linewidth]{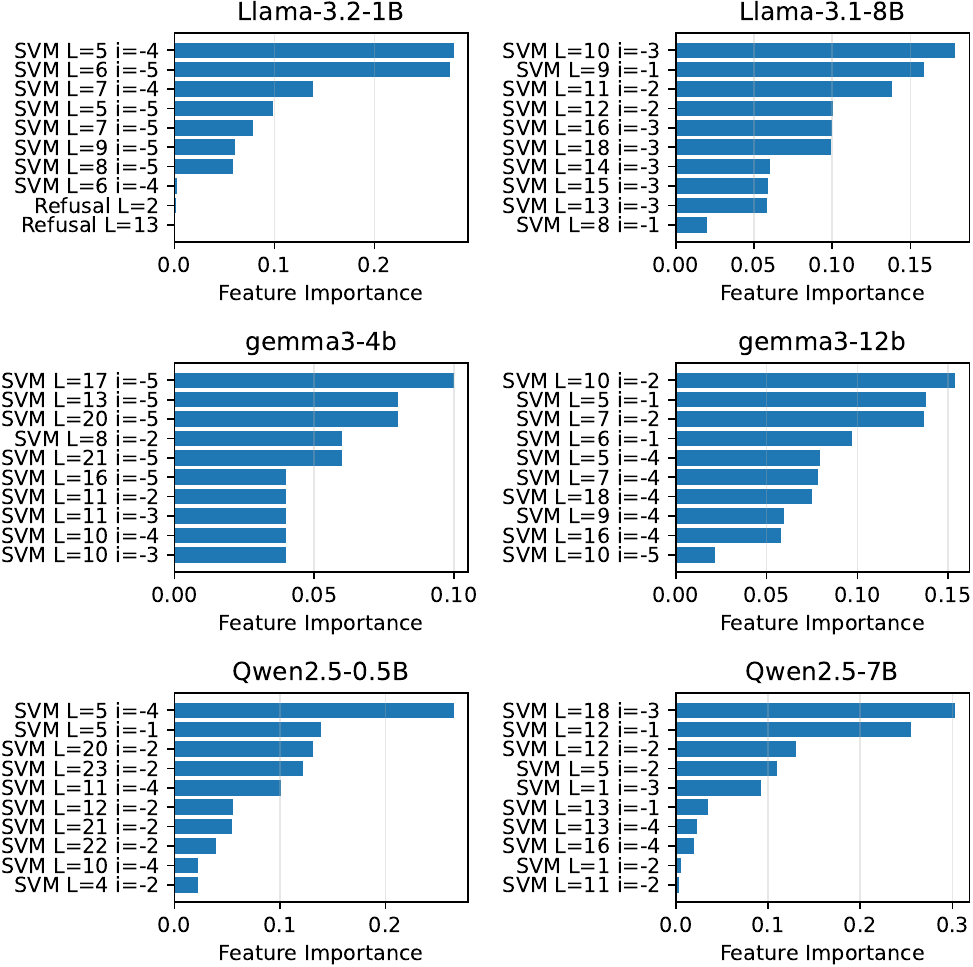}}
\caption{Top 10 feature importance of AlignTree's RandomForest for 2 models in each model family; Llama-3.2-1B-Instruct, Llama-3.2-8B-Instruct, gemma3-4b-it, gemma3-12b-it, Qwen2.5-0.5B-Instruct, Qwen2.5-7B-Instruct. }
\label{fig: feature_importance}

\end{figure*}

\clearpage
\clearpage

\section{Additional results}
\label{sec:full_results}
This section presents additional results, including extensive benchmark evaluations and a further ablation study of the AlignTree defense method.

\subsection{ASR}
\label{section:asr_results}
We provide additional ASR results for all datasets across all defense strategies. The results for Llama-3 Instruct models are shown in Table~\ref{tab:llama_asr}, for Gemma-3 Instruct models in Table~\ref{tab:gemma_asr}, and for Qwen-2.5 Instruct models in Table~\ref{tab:qwen_asr}.

\begin{table*}[t]
\centering
\begin{small}
\scalebox{0.97}{
\begin{tabular}{llcccccccc}
\toprule
Model & Strategy & \multicolumn{2}{c}{MalwareGen} & \multicolumn{2}{c}{PromptInject}  & \multicolumn{2}{c}{PAIR} & \multicolumn{2}{c}{AutoDAN} \\\cmidrule(lr){3-4} \cmidrule(lr){5-6} \cmidrule(lr){7-8} \cmidrule(lr){9-10}
& & ASR & Time & ASR & Time & ASR & Time & ASR & Time  \\
\midrule
Llama & Baseline & 32.0 & 14.17s & 36.0 & 10.25s & 12.0 & \textbf{17.02}s & 1.0 & 4.79s \\ 
-3.2-1B &  AutoDefense & \textbf{3.0} & 825.07s & \textbf{2.0} & 469.17s & 8.0 & 653.75s & 1.0 & 502.02s \\ 
-Instruct & SelfDefense-Input & 28.0 & 26.56s & 24.0 & 15.89s & \textbf{5.0} & 35.63s & 1.0 & 18.62s \\ 
 & SelfDefense & 23.0 & 26.33s & 24.0 & 16.74s & 6.0 & 28.0s & 1.0 & 19.43s \\ 
 & PerplexityDefense & 26.0 & 14.78s & 35.0 & 10.92s & 13.0 & 17.81s & 1.0 & 5.91s \\ 
 & SmoothLLM & 30.0 & 146.36s & 13.0 & 117.05s & 10.0 & 122.68s & \textbf{0} & 75.62s \\ 
 & AlignTree (\textbf{Ours}) & 20.0 & \textbf{11.67}s & 21.0 & \textbf{5.75}s & 6.0 & 18.64s & \textbf{0} & \textbf{1.05}s \\ 
\midrule 
Llama & Baseline & 16.0 & 54.37s & 28.0 & 23.27s & 19.0 & 59.82s & 1.0 & 10.7s \\ 
-3.2-3B &  AutoDefense & \textbf{1.0} & 566.02s & \textbf{2.0} & 489.71s & 17.0 & 937.0s & \textbf{0} & 247.46s \\ 
-Instruct & SelfDefense-Input & 13.0 & 53.28s & 21.0 & \textbf{14.46}s & 17.0 & 68.19s & 1.0 & 19.33s \\ 
 & SelfDefense & 15.0 & 48.38s & 22.0 & 15.36s & 18.0 & \textbf{51.29}s & 1.0 & 17.61s \\ 
 & PerplexityDefense & 15.0 & 54.5s & 26.0 & 24.13s & 19.0 & 58.45s & 1.0 & 12.16s \\ 
 & SmoothLLM & 15.0 & 263.22s & 20.0 & 203.48s & 11.0 & 214.34s & \textbf{0} & 148.88s \\ 
 & AlignTree (\textbf{Ours}) & 12.0 & \textbf{47.53}s & 17.0 & 14.88s & \textbf{10.0} & 53.54s & \textbf{0} & \textbf{5.99}s \\ 
\midrule 
Llama & Baseline & 9.0 & 174.53s & 43.0 & 51.97s & 14.0 & 274.71s & \textbf{0} & 29.1s \\ 
-3.1-8B &  AutoDefense & \textbf{5.0} & 553.46s & \textbf{0} & 361.85s & 16.0 & 837.72s & \textbf{0} & 299.65s \\ 
-Instruct & SelfDefense-Input & 8.0 & 247.66s & 32.0 & 123.06s & \textbf{8.0} & 333.96s & \textbf{0} & 138.42s \\ 
 & SelfDefense & 8.0 & 202.17s & 28.0 & 95.57s & \textbf{8.0} & 253.02s & \textbf{0} & 98.06s \\ 
 & PerplexityDefense & 8.0 & 178.81s & 42.0 & 54.55s & 15.0 & 270.72s & \textbf{0} & 32.22s \\ 
 & SmoothLLM & 8.0 & 505.02s & 37.0 & 383.04s & 13.0 & 525.58s & \textbf{0} & 271.95s \\ 
 & AlignTree (\textbf{Ours}) & \textbf{5.0} & \textbf{87.37}s & 18.0 & \textbf{34.2}s & 9.0 & \textbf{128.94}s & \textbf{0} & \textbf{5.8}s \\ 
\bottomrule
\end{tabular}
}
\end{small}
\caption{Additional results for Llama Instruct models on harmful benchmarks. The Attack Success Rate for each dataset, with the execution time for all defense strategies across all datasets.}
\label{tab:llama_asr}
\end{table*}

\begin{table*}[t]
\centering
\begin{small}
\scalebox{0.97}{
\begin{tabular}{llcccccccc}
\toprule
Model & Strategy & \multicolumn{2}{c}{MalwareGen} & \multicolumn{2}{c}{PromptInject}  & \multicolumn{2}{c}{PAIR} & \multicolumn{2}{c}{AutoDAN} \\\cmidrule(lr){3-4} \cmidrule(lr){5-6} \cmidrule(lr){7-8} \cmidrule(lr){9-10}
& & ASR & Time & ASR & Time & ASR & Time & ASR & Time  \\
\midrule
gemma & Baseline & 37.0 & \textbf{36.86}s & 39.0 & 37.08s & 35.0 & 37.65s & 8.0 & 36.78s \\ 
-3-1b &  AutoDefense & 19.0 & 1136.23s & 4.0 & 741.54s & 22.0 & 1107.78s & 1.0 & 1152.82s \\ 
-it & SelfDefense-Input & \textbf{1.0} & 80.95s & 5.0 & \textbf{22.49}s & \textbf{0} & \textbf{19.22}s & \textbf{0} & 20.27s \\ 
 & SelfDefense & 2.0 & 79.47s & \textbf{2.0} & 34.2s & \textbf{0} & 44.86s & \textbf{0} & \textbf{17.32}s \\ 
 & PerplexityDefense & 21.0 & 53.16s & 33.0 & 52.41s & 34.0 & 56.69s & 6.0 & 46.51s \\ 
 & SmoothLLM & 35.0 & 404.96s & 35.0 & 353.55s & 23.0 & 378.94s & 2.0 & 393.65s \\ 
 & AlignTree (\textbf{Ours}) & 30.0 & 42.23s & 40.0 & 52.7s & 23.0 & 38.85s & 1.0 & 36.76s \\ 
\midrule 
gemma & Baseline & 31.0 & 127.2s & 54.0 & 109.94s & 36.0 & 116.19s & 16.0 & 123.17s \\ 
-3-4b &  AutoDefense & 5.0 & 1166.97s & 4.0 & 299.04s & 18.0 & 1308.45s & 1.0 & 1350.8s \\ 
-it & SelfDefense-Input & \textbf{4.0} & 130.5s & \textbf{0} & 28.75s & \textbf{12.0} & 196.13s & \textbf{0} & 29.06s \\ 
 & SelfDefense & \textbf{4.0} & 131.11s & \textbf{0} & \textbf{26.57}s & \textbf{12.0} & 134.0s & \textbf{0} & 27.71s \\ 
 & PerplexityDefense & 16.0 & \textbf{94.76}s & 47.0 & 138.73s & 31.0 & 146.75s & 15.0 & 133.15s \\ 
 & SmoothLLM & 29.0 & 523.68s & 55.0 & 532.96s & 34.0 & 532.66s & 10.0 & 546.91s \\ 
 & AlignTree (\textbf{Ours}) & 29.0 & 105.86s & 55.0 & 108.92s & 13.0 & \textbf{114.08}s & \textbf{0} & \textbf{4.24}s \\ 
\midrule 
gemma & Baseline & 24.0 & 963.31s & 50.0 & 959.33s & 36.0 & 942.08s & 6.0 & 954.95s \\ 
-3-12b &  AutoDefense & \textbf{7.0} & 1679.09s & \textbf{5.0} & 961.3s & 19.0 & 1885.07s & \textbf{1.0} & 1739.91s \\ 
-it & SelfDefense-Input & 23.0 & 845.26s & 35.0 & 1178.87s & 28.0 & 1131.6s & 3.0 & 383.83s \\ 
 & SelfDefense & 18.0 & 1355.14s & \textbf{5.0} & 1292.2s & 33.0 & 1309.25s & 4.0 & 1370.69s \\ 
 & PerplexityDefense & 16.0 & 965.51s & 52.0 & 982.62s & 35.0 & 995.25s & 5.0 & 982.46s \\ 
 & SmoothLLM & 25.0 & 1865.47s & 55.0 & 1874.88s & 37.0 & 1894.6s & 7.0 & 1876.13s \\ 
 & AlignTree (\textbf{Ours}) & 10.0 & \textbf{591.11}s & 40.0 & \textbf{717.22}s & \textbf{10.0} & \textbf{329.52}s & \textbf{1.0} & \textbf{127.92}s \\ 
\bottomrule
\end{tabular}
}
\end{small}
\caption{Additional results for Gemma Instruct models on harmful benchmarks. The Attack Success Rate for each dataset, with the execution time for all defense strategies across all datasets. }
\label{tab:gemma_asr}
\end{table*}

\subsection{Refusal}
\label{section:refusal_results_appendix}
We present the refusal results for all LLama-3 Instruct models and datasets in Table~\ref{tab:llama_refusal}, Gemma3 Instruct models and datasets in Table~\ref{tab:gemma_refusal}, and Qwen2.5 Instruct models and datasets in Table~\ref{tab:qwen_refusal}.

\begin{table*}[t]
\centering
\begin{small}
\scalebox{0.95}{
\begin{tabular}{llcccccccc}
\toprule
Model & Strategy & \multicolumn{2}{c}{MalwareGen} & \multicolumn{2}{c}{PromptInject}  & \multicolumn{2}{c}{PAIR} & \multicolumn{2}{c}{AutoDAN} \\\cmidrule(lr){3-4} \cmidrule(lr){5-6} \cmidrule(lr){7-8} \cmidrule(lr){9-10}
& & ASR & Time & ASR & Time & ASR & Time & ASR & Time  \\
\midrule
Qwen2.5 & Baseline & 91.0 & 19.12s & 50.0 & 18.4s & 51.0 & 19.57s & 48.0 & 18.56s \\ 
-0.5B &  AutoDefense & 5.0 & 1027.52s & \textbf{0} & 343.8s & 13.0 & 915.98s & \textbf{0} & 781.04s \\ 
-Instruct & SelfDefense-Input & 43.0 & 38.77s & 13.0 & 27.91s & 8.0 & 39.01s & 17.0 & 38.86s \\ 
 & SelfDefense & 42.0 & 38.56s & 16.0 & \textbf{13.09}s & 11.0 & 37.26s & 13.0 & 38.36s \\ 
 & PerplexityDefense & 84.0 & 21.2s & 50.0 & 21.03s & 50.0 & 21.66s & 47.0 & 21.02s \\ 
 & SmoothLLM & 77.0 & 210.79s & 43.0 & 209.75s & 49.0 & 213.53s & 44.0 & 212.22s \\ 
 & AlignTree (\textbf{Ours}) & \textbf{4.0} & \textbf{19.01}s & 41.0 & 24.8s & \textbf{6.0} & \textbf{18.46}s & \textbf{0} & \textbf{5.24}s \\ 
\midrule 
Qwen2.5 & Baseline & 32.0 & 65.2s & 56.0 & 33.14s & 37.0 & 65.69s & 29.0 & 65.01s \\ 
-3B &  AutoDefense & 3.0 & 983.72s & \textbf{1.0} & 369.39s & 21.0 & 911.47s & 3.0 & 946.27s \\ 
-Instruct & SelfDefense-Input & 4.0 & 71.45s & 32.0 & 41.19s & 21.0 & 86.78s & 12.0 & 89.78s \\ 
 & SelfDefense & 9.0 & 76.85s & 29.0 & 40.19s & 24.0 & 79.74s & 13.0 & 81.35s \\ 
 & PerplexityDefense & 26.0 & 65.82s & 40.0 & 29.97s & 37.0 & 66.17s & 29.0 & 65.41s \\ 
 & SmoothLLM & 15.0 & 304.3s & 53.0 & 249.86s & 25.0 & 320.95s & 14.0 & 329.48s \\ 
 & AlignTree (\textbf{Ours}) & \textbf{1.0} & \textbf{29.04}s & 12.0 & \textbf{10.47}s & \textbf{14.0} & \textbf{63.51}s & \textbf{0} & \textbf{12.65}s \\ 
\midrule 
Qwen2.5 & Baseline & 43.0 & 284.2s & 58.0 & 81.13s & 37.0 & 297.47s & 14.0 & 271.41s \\ 
-7B &  AutoDefense & \textbf{3.0} & 983.72s & \textbf{1.0} & 369.39s & 21.0 & 911.47s & 3.0 & 946.27s \\ 
-Instruct & SelfDefense-Input & 40.0 & 468.83s & 55.0 & 150.4s & 36.0 & 376.8s & 14.0 & 323.25s \\ 
 & SelfDefense & 41.0 & 348.93s & 53.0 & 156.63s & 36.0 & 389.38s & 14.0 & 340.84s \\ 
 & PerplexityDefense & 34.0 & 337.79s & 44.0 & 71.91s & 38.0 & 320.17s & 15.0 & 272.2s \\ 
 & SmoothLLM & 43.0 & 601.36s & 63.0 & 329.86s & 36.0 & 649.49s & 14.0 & 591.41s \\ 
 & AlignTree (\textbf{Ours}) & 6.0 & \textbf{77.19}s & \textbf{1.0} & \textbf{3.93}s & \textbf{14.0} & \textbf{154.06}s & \textbf{0} & \textbf{1.48}s \\
\bottomrule
\end{tabular}
}
\end{small}
\caption{Additional results for Qwen Instruct models on harmful benchmarks. The Attack Success Rate for each dataset, with the execution time for all defense strategies across all datasets.}
\label{tab:qwen_asr}
\end{table*}

\begin{table*}
\centering
\begin{small}
\scalebox{0.95}{
\begin{tabular}{llcccccccc}
\toprule
Model & Strategy & \multicolumn{2}{c}{PIQA} & \multicolumn{2}{c}{OpenbookQA}  & \multicolumn{2}{c}{SocialIQA} & \multicolumn{2}{c}{ARC-Challenge}  \\\cmidrule(lr){3-4} \cmidrule(lr){5-6} \cmidrule(lr){7-8} \cmidrule(lr){9-10}
& & Refusal & Time & Refusal & Time & Refusal & Time & Refusal & Time  \\
\midrule
Llama & Baseline & \textbf{3.0} & 1.29s & \textbf{0} & 5.13s & \textbf{3.0} & \textbf{2.9}s & \textbf{0} & 6.1s \\ 
-3.2-1B &  AutoDefense & 6.0 & 76.62s & 10.0 & 381.94s & 6.0 & 210.6s & 12.0 & 627.91s \\ 
-Instruct & SelfDefense-Input & 57.0 & 7.64s & 38.0 & 12.02s & 36.0 & 9.74s & 49.0 & 15.9s \\ 
 & SelfDefense & 58.0 & 8.59s & 30.0 & 12.91s & 35.0 & 10.04s & 39.0 & 15.12s \\ 
 & PerplexityDefense & \textbf{3.0} & 2.29s & \textbf{0} & 6.16s & \textbf{3.0} & 3.96s & \textbf{0} & 7.45s \\ 
 & SmoothLLM & 21.0 & 132.1s & 12.0 & 132.04s & 26.0 & 126.87s & \textbf{0} & 137.16s \\ 
 & AlignTree (\textbf{Ours}) & \textbf{3.0} & \textbf{0.7}s & \textbf{0} & \textbf{4.56}s & \textbf{3.0} & 2.96s & \textbf{0} & \textbf{5.28}s \\ 
\midrule 
Llama & Baseline & \textbf{10.0} & \textbf{4.0}s & \textbf{1.0} & 4.47s & \textbf{10.0} & \textbf{3.51}s & \textbf{0} & \textbf{8.94}s \\ 
-3.2-3B &  AutoDefense & 53.0 & 255.5s & 6.0 & 155.17s & 15.0 & 331.77s & 4.0 & 325.72s \\ 
-Instruct & SelfDefense-Input & 36.0 & 5.07s & 21.0 & 6.13s & 18.0 & 4.6s & 16.0 & 10.23s \\ 
 & SelfDefense & 35.0 & 4.54s & 12.0 & \textbf{4.28}s & 21.0 & 3.61s & 12.0 & 9.12s \\ 
 & PerplexityDefense & \textbf{10.0} & 6.06s & \textbf{1.0} & 6.12s & \textbf{10.0} & 5.1s & \textbf{0} & 10.83s \\ 
 & SmoothLLM & 14.0 & 214.38s & 6.0 & 214.28s & 23.0 & 208.42s & 1.0 & 222.11s \\ 
 & AlignTree (\textbf{Ours}) & \textbf{10.0} & 4.86s & \textbf{1.0} & 12.24s & \textbf{10.0} & 5.05s & \textbf{0} & 15.28s \\ 
\midrule 
Llama & Baseline & 2.0 & \textbf{15.59}s & \textbf{0} & 32.53s & \textbf{5.0} & 12.52s & \textbf{0} & 56.78s \\ 
-3.1-8B &  AutoDefense & 2.0 & 86.66s & 1.0 & 272.5s & 7.0 & 238.2s & 4.0 & 292.15s \\ 
-Instruct & SelfDefense-Input & 55.0 & 87.29s & 47.0 & 138.26s & 30.0 & 79.5s & 52.0 & 181.8s \\ 
 & SelfDefense & 51.0 & 95.86s & 55.0 & 85.34s & 34.0 & 81.92s & 49.0 & 104.44s \\ 
 & PerplexityDefense & 2.0 & 18.93s & \textbf{0} & 35.58s & \textbf{5.0} & 15.7s & \textbf{0} & 59.86s \\ 
 & SmoothLLM & 2.0 & 407.79s & \textbf{0} & 414.91s & \textbf{5.0} & 399.89s & \textbf{0} & 456.06s \\ 
 & AlignTree (\textbf{Ours}) & \textbf{1.0} & 21.88s & \textbf{0} & \textbf{25.09}s & \textbf{5.0} & \textbf{9.02}s & \textbf{0} & \textbf{37.44}s \\ 
\bottomrule
\end{tabular}
}
\end{small}
\caption{Additional results for Llama Instruct models on harmless, commonsense benchmarks. These experiments measure excessive refusal and time efficiency for harmless prompts.}
\label{tab:llama_refusal}
\end{table*}

\begin{table*}
\centering
\begin{small}
\scalebox{0.97}{
\begin{tabular}{llcccccccc}
\toprule
Model & Strategy & \multicolumn{2}{c}{PIQA} & \multicolumn{2}{c}{OpenbookQA}  & \multicolumn{2}{c}{SocialIQA} & \multicolumn{2}{c}{ARC-Challenge}  \\\cmidrule(lr){3-4} \cmidrule(lr){5-6} \cmidrule(lr){7-8} \cmidrule(lr){9-10}
& & Refusal & Time & Refusal & Time & Refusal & Time & Refusal & Time  \\
\midrule
gemma & Baseline & \textbf{0} & \textbf{2.12}s & \textbf{0} & \textbf{2.35}s & \textbf{0} & \textbf{3.89}s & \textbf{0} & 5.41s \\ 
-3-1b &  AutoDefense & 32.0 & 186.99s & 2.0 & 535.29s & 2.0 & 395.68s & 2.0 & 474.96s \\ 
-it & SelfDefense-Input & 8.0 & 16.88s & 15.0 & 16.17s & \textbf{0} & 18.52s & 16.0 & 22.51s \\ 
 & SelfDefense & 7.0 & 14.82s & 20.0 & 13.96s & 3.0 & 18.55s & 16.0 & 18.94s \\ 
 & PerplexityDefense & \textbf{0} & 22.31s & \textbf{0} & 8.82s & \textbf{0} & 10.6s & \textbf{0} & 27.05s \\ 
 & SmoothLLM & \textbf{0} & 273.27s & \textbf{0} & 295.69s & \textbf{0} & 294.47s & \textbf{0} & 329.92s \\ 
 & AlignTree (\textbf{Ours}) & \textbf{0} & 2.69s & \textbf{0} & 2.94s & \textbf{0} & 5.64s & \textbf{0} & \textbf{3.42}s \\ 
\midrule 
gemma & Baseline & \textbf{0} & \textbf{112.36}s & \textbf{0} & \textbf{109.69}s & \textbf{0} & \textbf{109.25}s & \textbf{0} & \textbf{110.08}s \\ 
-3-4b &  AutoDefense & \textbf{0} & 225.66s & 3.0 & 206.92s & 10.0 & 248.76s & 1.0 & 248.14s \\ 
-it & SelfDefense-Input & 2.0 & 200.86s & \textbf{0} & 150.37s & 3.0 & 253.44s & 1.0 & 168.44s \\ 
 & SelfDefense & 2.0 & 133.81s & \textbf{0} & 134.87s & 3.0 & 133.5s & 1.0 & 136.61s \\ 
 & PerplexityDefense & \textbf{0} & 140.85s & \textbf{0} & 130.43s & \textbf{0} & 131.23s & \textbf{0} & 134.77s \\ 
 & SmoothLLM & \textbf{0} & 538.26s & \textbf{0} & 532.74s & \textbf{0} & 534.65s & \textbf{0} & 538.81s \\ 
 & AlignTree (\textbf{Ours}) & \textbf{0} & 120.69s & \textbf{0} & 127.04s & \textbf{0} & 122.42s & \textbf{0} & 126.4s \\ 
\midrule 
gemma & Baseline & \textbf{0} & \textbf{955.92}s & \textbf{0} & \textbf{955.25}s & \textbf{0} & \textbf{961.94}s & \textbf{0} & \textbf{972.13}s \\ 
-3-12b &  AutoDefense & \textbf{0} & 1097.71s & 7.0 & 1203.54s & \textbf{0} & 2057.22s & 2.0 & 2017.82s \\ 
-it & SelfDefense-Input & 1.0 & 1182.44s & \textbf{0} & 1181.36s & 1.0 & 1179.36s & \textbf{0} & 1193.1s \\ 
 & SelfDefense & 2.0 & 1319.66s & 17.0 & 1186.86s & 58.0 & 1190.3s & 11.0 & 1188.49s \\ 
 & PerplexityDefense & \textbf{0} & 996.26s & \textbf{0} & 993.79s & \textbf{0} & 989.72s & \textbf{0} & 1033.18s \\ 
 & SmoothLLM & \textbf{0} & 1926.51s & \textbf{0} & 1935.92s & \textbf{0} & 1890.58s & \textbf{0} & 1913.35s \\ 
 & AlignTree (\textbf{Ours}) & \textbf{0} & 988.4s & \textbf{0} & 975.01s & \textbf{0} & 973.6s & \textbf{0} & 978.93s \\ 
\bottomrule
\end{tabular}
}
\end{small}
\caption{Additional results for Gemma Instruct models on harmless, commonsense benchmarks. These experiments measure excessive refusal and time efficiency for harmless prompts. }
\label{tab:gemma_refusal}
\end{table*}

\begin{table*}
\centering
\begin{small}
\scalebox{0.97}{
\begin{tabular}{llcccccccc}
\toprule
Model & Strategy & \multicolumn{2}{c}{PIQA} & \multicolumn{2}{c}{OpenbookQA}  & \multicolumn{2}{c}{SocialIQA} & \multicolumn{2}{c}{ARC-Challenge}  \\\cmidrule(lr){3-4} \cmidrule(lr){5-6} \cmidrule(lr){7-8} \cmidrule(lr){9-10}
& & Refusal & Time & Refusal & Time & Refusal & Time & Refusal & Time  \\
\midrule
Qwen2.5 & Baseline & \textbf{0} & \textbf{0.43}s & \textbf{0} & \textbf{0.7}s & \textbf{0} & \textbf{2.09}s & \textbf{0} & \textbf{0.7}s \\ 
-0.5B &  AutoDefense & \textbf{0} & 50.54s & 6.0 & 154.79s & 3.0 & 170.17s & 8.0 & 151.56s \\ 
-Instruct & SelfDefense-Input & 80.0 & 10.31s & 35.0 & 10.1s & 33.0 & 10.26s & 46.0 & 11.82s \\ 
 & SelfDefense & 72.0 & 9.97s & 37.0 & 10.02s & 37.0 & 12.27s & 41.0 & 10.19s \\ 
 & PerplexityDefense & \textbf{0} & 2.21s & \textbf{0} & 2.24s & \textbf{0} & 3.84s & \textbf{0} & 2.39s \\ 
 & SmoothLLM & \textbf{0} & 196.47s & \textbf{0} & 187.81s & \textbf{0} & 189.37s & \textbf{0} & 188.72s \\ 
 & AlignTree (\textbf{Ours}) & \textbf{0} & 3.16s & \textbf{0} & 1.16s & \textbf{0} & 4.86s & \textbf{0} & 1.27s \\
\midrule 
Qwen & Baseline & \textbf{0} & \textbf{3.21}s & \textbf{0} & \textbf{0.89}s & \textbf{0} & \textbf{4.18}s & \textbf{0} & \textbf{1.0}s \\ 
2.5-3B &  AutoDefense & \textbf{0} & 67.09s & \textbf{0} & 73.38s & 1.0 & 74.67s & \textbf{0} & 59.11s \\ 
-Instruct & SelfDefense-Input & 3.0 & 18.71s & \textbf{0} & 19.19s & \textbf{0} & 19.7s & \textbf{0} & 18.17s \\ 
 & SelfDefense & \textbf{0} & 15.73s & \textbf{0} & 17.7s & \textbf{0} & 19.18s & \textbf{0} & 17.43s \\ 
 & PerplexityDefense & \textbf{0} & 5.62s & \textbf{0} & 3.17s & \textbf{0} & 5.73s & \textbf{0} & 3.25s \\ 
 & SmoothLLM & \textbf{0} & 257.3s & \textbf{0} & 251.21s & \textbf{0} & 249.99s & \textbf{0} & 251.32s \\ 
 & AlignTree (\textbf{Ours}) & \textbf{0} & 4.47s & \textbf{0} & 2.82s & \textbf{0} & 4.84s & \textbf{0} & 3.95s \\ 
\midrule 
Qwen & Baseline & \textbf{0} & \textbf{10.26}s & \textbf{0} & \textbf{21.42}s & \textbf{0} & 9.42s & \textbf{0} & 41.98s \\ 
2.5-7B &  AutoDefense & \textbf{0} & 96.62s & \textbf{0} & 111.04s & \textbf{0} & 99.26s & \textbf{0} & 156.18s \\ 
-Instruct & SelfDefense-Input & \textbf{0} & 96.25s & \textbf{0} & 100.68s & \textbf{0} & 113.66s & \textbf{0} & 151.98s \\ 
 & SelfDefense & \textbf{0} & 87.89s & \textbf{0} & 76.81s & \textbf{0} & 71.91s & \textbf{0} & 99.92s \\ 
 & PerplexityDefense & \textbf{0} & 17.32s & \textbf{0} & 24.25s & \textbf{0} & 12.33s & \textbf{0} & 44.86s \\ 
 & SmoothLLM & \textbf{0} & 327.57s & \textbf{0} & 305.28s & \textbf{0} & 265.46s & \textbf{0} & 355.54s \\ 
 & AlignTree (\textbf{Ours}) & \textbf{0} & 11.57s & \textbf{0} & 29.54s & \textbf{0} & \textbf{9.2}s & \textbf{0} & \textbf{33.47}s \\
\bottomrule
\end{tabular}
}
\end{small}
\caption{Additional results for Qwen Instruct models on harmless, commonsense benchmarks. These experiments measure excessive refusal and time efficiency for harmless prompts. }
\label{tab:qwen_refusal}
\end{table*}

\subsection{Further ablation results}
\label{appx:ablation}
 In this section, we provide ablation results for all nine models across eight datasets. The ASR results, presented in Table~\ref{tab:full_asr_ablation_study} and refusal results presented in Table~\ref{tab:full_refusal_ablation_study}, depict the same trends as the partial table. 

\begin{table*}[h!]
\centering
\begin{small}
\begin{tabular}{llcccccccc}
\toprule
Model & Strategy & \multicolumn{2}{c}{MalwareGen} & \multicolumn{2}{c}{PromptInject}  & \multicolumn{2}{c}{PAIR} & \multicolumn{2}{c}{AutoDAN} \\\cmidrule(lr){3-4} \cmidrule(lr){5-6} \cmidrule(lr){7-8} \cmidrule(lr){9-10}
& & ASR & Time & ASR & Time & ASR & Time & ASR & Time  \\
\midrule
Llama-3.2-1B & RefusalClassifier & \textbf{1.0} & 17.46s & \textbf{10.0} & 5.93s & 3.0 & 19.57s & \textbf{0} & 0.55s \\ 
-Instruct & SVMClassifier & 29.0 & 12.19s & 35.0 & 7.56s & 10.0 & 11.79s & \textbf{0} & \textbf{0.52}s \\ 
 & MultiRefusalsClassifier & 14.0 & 11.56s & 28.0 & 5.94s & 9.0 & 11.84s & \textbf{0} & 0.83s \\ 
 & AlignTreeLinear & 20.0 & \textbf{11.34}s & 21.0 & \textbf{3.6}s & \textbf{2.0} & \textbf{11.17}s & \textbf{0} & 1.25s \\ 
 & AlignTree & 20.0 & 11.67s & 21.0 & 5.75s & 6.0 & 18.64s & \textbf{0} & 1.05s \\ 
\midrule 
Llama-3.2-3B & RefusalClassifier & \textbf{3.0} & 44.14s & \textbf{8.0} & 22.45s & \textbf{8.0} & 63.55s & \textbf{0} & 4.8s \\ 
-Instruct & SVMClassifier & 5.0 & \textbf{35.22}s & 15.0 & \textbf{10.86}s & 9.0 & \textbf{46.46}s & 1.0 & 4.74s \\ 
 & MultiRefusalsClassifier & 10.0 & 53.35s & 24.0 & 18.42s & 17.0 & 48.73s & \textbf{0} & \textbf{1.88}s \\ 
 & AlignTreeLinear & 4.0 & 45.24s & 19.0 & 14.4s & 11.0 & 47.67s & \textbf{0} & 3.24s \\ 
 & AlignTree & 12.0 & 47.53s & 17.0 & 14.88s & 10.0 & 53.54s & \textbf{0} & 5.99s \\ 
\midrule 
Llama-3.1-8B & RefusalClassifier & 5.0 & 145.54s & 44.0 & 57.68s & 11.0 & 230.5s & \textbf{0} & 12.64s \\ 
-Instruct & SVMClassifier & \textbf{2.0} & 66.58s & 20.0 & 42.02s & 4.0 & 66.3s & \textbf{0} & 1.75s \\ 
 & MultiRefusalsClassifier & 4.0 & \textbf{62.49}s & 32.0 & \textbf{31.67}s & \textbf{1.0} & \textbf{21.7}s & \textbf{0} & \textbf{0.8}s \\ 
 & AlignTreeLinear & 7.0 & 101.99s & \textbf{18.0} & 40.61s & 9.0 & 122.13s & \textbf{0} & 6.33s \\ 
 & AlignTree & 5.0 & 87.37s & \textbf{18.0} & 34.2s & 9.0 & 128.94s & \textbf{0} & 5.8s \\
\midrule 
gemma-3-1b & RefusalClassifier & 28.0 & 74.28s & 40.0 & 61.04s & 31.0 & 57.78s & 6.0 & 311.39s \\ 
-it & SVMClassifier & \textbf{23.0} & 41.77s & \textbf{30.0} & 49.07s & \textbf{6.0} & \textbf{33.5}s & \textbf{0} & \textbf{1.72}s \\ 
 & MultiRefusalsClassifier & 40.0 & 39.3s & 34.0 & \textbf{40.11}s & 32.0 & 40.68s & 6.0 & 37.31s \\ 
 & AlignTreeLinear & 33.0 & \textbf{35.61}s & \textbf{30.0} & 49.05s & 26.0 & 40.3s & 5.0 & 32.8s \\ 
 & AlignTree & 30.0 & 42.23s & 40.0 & 52.7s & 23.0 & 38.85s & 1.0 & 36.76s \\ 
\midrule 
gemma-3-4b & RefusalClassifier & 31.0 & 158.01s & 54.0 & 189.08s & 26.0 & 196.93s & 6.0 & 185.28s \\ 
-it & SVMClassifier & \textbf{6.0} & 133.65s & \textbf{10.0} & 123.63s & \textbf{1.0} & 100.2s & \textbf{0} & 6.59s \\ 
 & MultiRefusalsClassifier & 14.0 & 108.14s & 53.0 & 114.94s & 16.0 & 120.92s & 2.0 & 85.4s \\ 
 & AlignTreeLinear & 10.0 & 151.25s & 32.0 & 121.67s & \textbf{1.0} & \textbf{41.7}s & \textbf{0} & 37.31s \\ 
 & AlignTree & 29.0 & \textbf{105.86}s & 55.0 & \textbf{108.92}s & 13.0 & 114.08s & \textbf{0} & \textbf{4.24}s \\ 
\midrule 
gemma-3-12b & RefusalClassifier & 21.0 & 619.41s & 54.0 & 957.3s & 22.0 & 910.53s & 1.0 & 57.31s \\ 
-it & SVMClassifier & 26.0 & 738.11s & 37.0 & 708.4s & \textbf{5.0} & \textbf{175.27}s & \textbf{0} & 33.83s \\ 
 & MultiRefusalsClassifier & 25.0 & 509.97s & 53.0 & \textbf{602.4}s & 26.0 & 523.65s & \textbf{0} & 37.47s \\ 
 & AlignTreeLinear & \textbf{8.0} & \textbf{496.35}s & \textbf{29.0} & 800.97s & 8.0 & 350.06s & \textbf{0} & \textbf{4.78}s \\ 
 & AlignTree & 10.0 & 591.11s & 40.0 & 717.22s & 10.0 & 329.52s & 1.0 & 127.92s \\
\midrule 
Qwen2.5-0.5B & RefusalClassifier & 89.0 & 27.18s & 52.0 & 27.34s & 50.0 & 27.65s & 48.0 & 27.46s \\ 
-Instruct & SVMClassifier & 33.0 & 21.85s & 46.0 & 27.05s & \textbf{4.0} & 20.61s & \textbf{0} & \textbf{0.72}s \\ 
 & MultiRefusalsClassifier & 29.0 & \textbf{17.45}s & 53.0 & 18.32s & 36.0 & \textbf{17.97}s & 44.0 & 18.38s \\ 
 & AlignTreeLinear & 61.0 & 22.67s & 43.0 & \textbf{16.57}s & 23.0 & 20.04s & 7.0 & 19.74s \\ 
 & AlignTree & \textbf{4.0} & 19.01s & \textbf{41.0} & 24.8s & 6.0 & 18.46s & \textbf{0} & 5.24s \\ 
\midrule 
Qwen2.5-3B & RefusalClassifier & \textbf{0} & 11.78s & 54.0 & 167.59s & 9.0 & 94.41s & 7.0 & 60.49s \\ 
-Instruct & SVMClassifier & 4.0 & 35.72s & 52.0 & 43.72s & 6.0 & \textbf{50.58}s & \textbf{0} & 0.94s \\ 
 & MultiRefusalsClassifier & \textbf{0} & \textbf{0.43}s & 49.0 & 41.52s & 24.0 & 70.56s & 18.0 & 70.4s \\ 
 & AlignTreeLinear & 5.0 & 46.94s & 27.0 & 21.64s & \textbf{5.0} & 54.36s & \textbf{0} & \textbf{0.58}s \\ 
 & AlignTree & 1.0 & 29.04s & \textbf{12.0} & \textbf{10.47}s & 14.0 & 63.51s & \textbf{0} & 12.65s \\ 
\midrule 
Qwen2.5-7B & RefusalClassifier & 8.0 & 157.97s & 62.0 & 107.46s & 36.0 & 369.33s & 15.0 & 286.95s \\ 
-Instruct & SVMClassifier & 8.0 & 86.79s & 24.0 & 21.96s & \textbf{4.0} & \textbf{55.31}s & \textbf{0} & 1.78s \\ 
 & MultiRefusalsClassifier & 18.0 & 88.23s & 41.0 & 39.89s & 15.0 & 120.24s & 2.0 & 18.34s \\ 
 & AlignTreeLinear & 17.0 & 106.69s & 53.0 & 30.62s & 19.0 & 168.65s & 4.0 & 55.67s \\ 
 & AlignTree & \textbf{6.0} & \textbf{77.19}s & \textbf{1.0} & \textbf{3.93}s & 14.0 & 154.06s & \textbf{0} & \textbf{1.48}s \\
\bottomrule
\end{tabular}
\end{small}
\caption{Full ASR results for ablated defenses in AlignTree. We can observe the trend is similar to Table~\ref{tab:ablation_study}.}
\label{tab:full_asr_ablation_study}
\end{table*}

\begin{table*}[h!]
\centering
\begin{small}
\begin{tabular}{llcccccccc}
\toprule
Model & Strategy & \multicolumn{2}{c}{PIQA} & \multicolumn{2}{c}{OpenbookQA}  & \multicolumn{2}{c}{SocialIQA} & \multicolumn{2}{c}{ARC-Challenge} \\\cmidrule(lr){3-4} \cmidrule(lr){5-6} \cmidrule(lr){7-8} \cmidrule(lr){9-10}
& & Refusal & Time & Refusal & Time & Refusal & Time & Refusal & Time  \\
\midrule
Llama-3.2-1B & RefusalClassifier & \textbf{3.0} & 2.47s & \textbf{0} & 8.11s & 4.0 & 5.55s & \textbf{0} & 11.74s \\ 
-Instruct & SVMClassifier & \textbf{3.0} & 2.35s & \textbf{0} & 8.41s & \textbf{3.0} & 5.51s & \textbf{0} & 12.05s \\ 
 & MultiRefusalsClassifier & \textbf{3.0} & 1.26s & \textbf{0} & 4.63s & \textbf{3.0} & 2.87s & \textbf{0} & 5.42s \\ 
 & AlignTreeLinear & 4.0 & \textbf{0.65}s & \textbf{0} & 4.61s & \textbf{3.0} & \textbf{2.84}s & \textbf{0} & 5.4s \\ 
 & AlignTree & \textbf{3.0} & 0.7s & \textbf{0} & \textbf{4.56}s & \textbf{3.0} & 2.96s & \textbf{0} & \textbf{5.28}s \\ 
\midrule 
Llama-3.2-3B & RefusalClassifier & \textbf{10.0} & 8.53s & \textbf{1.0} & 7.91s & \textbf{10.0} & 6.04s & \textbf{0} & 13.19s \\ 
-Instruct & SVMClassifier & 11.0 & 6.03s & \textbf{1.0} & 42.38s & \textbf{10.0} & 35.42s & \textbf{0} & 53.34s \\ 
 & MultiRefusalsClassifier & \textbf{10.0} & 5.29s & \textbf{1.0} & 6.24s & \textbf{10.0} & 3.45s & \textbf{0} & 12.27s \\ 
 & AlignTreeLinear & 11.0 & \textbf{1.73}s & \textbf{1.0} & \textbf{6.06}s & \textbf{10.0} & \textbf{2.51}s & \textbf{0} & \textbf{8.67}s \\ 
 & AlignTree & \textbf{10.0} & 4.86s & \textbf{1.0} & 12.24s & \textbf{10.0} & 5.05s & \textbf{0} & 15.28s \\ 
\midrule 
Llama-3.1-8B & RefusalClassifier & 2.0 & 20.55s & \textbf{0} & 34.55s & \textbf{5.0} & 14.64s & \textbf{0} & 59.84s \\ 
-Instruct & SVMClassifier & 3.0 & 24.53s & \textbf{0} & 122.24s & \textbf{5.0} & 22.62s & \textbf{0} & 98.11s \\ 
 & MultiRefusalsClassifier & \textbf{1.0} & 11.66s & \textbf{0} & \textbf{20.02}s & \textbf{5.0} & 7.98s & \textbf{0} & \textbf{35.79}s \\ 
 & AlignTreeLinear & 3.0 & \textbf{8.21}s & \textbf{0} & 25.02s & \textbf{5.0} & \textbf{7.03}s & \textbf{0} & 43.44s \\ 
 & AlignTree & 3.0 & 13.29s & \textbf{0} & 35.86s & \textbf{5.0} & 13.37s & \textbf{0} & 60.47s \\ 
\midrule 
gemma-3-1b & RefusalClassifier & \textbf{0} & 17.13s & \textbf{0} & 11.69s & \textbf{0} & 15.04s & \textbf{0} & 350.36s \\ 
-it & SVMClassifier & \textbf{0} & 5.12s & \textbf{0} & 4.64s & \textbf{0} & 12.71s & \textbf{0} & 8.58s \\ 
 & MultiRefusalsClassifier & \textbf{0} & 3.21s & \textbf{0} & 3.08s & \textbf{0} & 6.66s & \textbf{0} & 4.71s \\ 
 & AlignTreeLinear & \textbf{0} & 5.02s & \textbf{0} & \textbf{2.51}s & \textbf{0} & 6.76s & \textbf{0} & 5.11s \\ 
 & AlignTree & \textbf{0} & \textbf{2.69}s & \textbf{0} & 2.94s & \textbf{0} & \textbf{5.64}s & \textbf{0} & \textbf{3.42}s \\ 
\midrule 
gemma-3-4b & RefusalClassifier & \textbf{0} & 190.45s & \textbf{0} & 158.67s & \textbf{0} & 133.95s & \textbf{0} & 162.7s \\ 
-it & SVMClassifier & 100.0 & \textbf{34.43}s & 100.0 & \textbf{35.24}s & 100.0 & \textbf{29.28}s & 100.0 & \textbf{28.13}s \\ 
 & MultiRefusalsClassifier & \textbf{0} & 118.85s & \textbf{0} & 114.18s & \textbf{0} & 117.22s & \textbf{0} & 117.45s \\ 
 & AlignTreeLinear & \textbf{0} & 139.14s & \textbf{0} & 185.25s & \textbf{0} & 168.52s & \textbf{0} & 175.44s \\ 
 & AlignTree & \textbf{0} & 120.69s & \textbf{0} & 127.04s & \textbf{0} & 122.42s & \textbf{0} & 126.4s \\ 
\midrule 
gemma-3-12b & RefusalClassifier & \textbf{0} & 981.92s & \textbf{0} & 960.61s & \textbf{0} & 953.05s & \textbf{0} & 983.69s \\ 
-it & SVMClassifier & 5.0 & 965.68s & \textbf{0} & 966.58s & 1.0 & 959.45s & \textbf{0} & 969.49s \\ 
 & MultiRefusalsClassifier & \textbf{0} & \textbf{640.47}s & \textbf{0} & 619.29s & \textbf{0} & \textbf{629.87}s & \textbf{0} & 606.22s \\ 
 & AlignTreeLinear & \textbf{0} & 761.13s & \textbf{0} & \textbf{596.65}s & \textbf{0} & 835.81s & \textbf{0} & \textbf{600.52}s \\ 
 & AlignTree & \textbf{0} & 988.4s & \textbf{0} & 975.01s & \textbf{0} & 973.6s & \textbf{0} & 978.93s \\ 
\midrule 
Qwen & RefusalClassifier & \textbf{0} & 1.44s & \textbf{0} & 1.65s & \textbf{0} & 3.57s & \textbf{0} & 1.79s \\ 
2.5-0.5B & SVMClassifier & \textbf{0} & 1.53s & \textbf{0} & 1.64s & \textbf{0} & 3.55s & \textbf{0} & 1.94s \\ 
 -Instruct & MultiRefusalsClassifier & \textbf{0} & \textbf{0.58}s & \textbf{0} & \textbf{0.81}s & \textbf{0} & \textbf{2.34}s & \textbf{0} & \textbf{0.89}s \\ 
 & AlignTreeLinear & \textbf{0} & 0.82s & \textbf{0} & 0.89s & \textbf{0} & 2.43s & \textbf{0} & 0.97s \\ 
 & AlignTree & \textbf{0} & 0.73s & \textbf{0} & 0.98s & \textbf{0} & 2.42s & \textbf{0} & 1.03s \\ 
\midrule 
Qwen & RefusalClassifier & \textbf{0} & 6.72s & \textbf{0} & 1.66s & \textbf{0} & 9.82s & \textbf{0} & 1.66s \\ 
2.5-3B & SVMClassifier & \textbf{0} & 114.3s & \textbf{0} & 164.26s & \textbf{0} & 629.9s & \textbf{0} & 203.46s \\ 
 -Instruct & MultiRefusalsClassifier & \textbf{0} & 2.34s & \textbf{0} & 1.78s & \textbf{0} & 4.68s & \textbf{0} & 2.04s \\ 
 & AlignTreeLinear & 1.0 & \textbf{1.65}s & \textbf{0} & \textbf{1.43}s & \textbf{0} & \textbf{4.02}s & \textbf{0} & \textbf{1.51}s \\ 
 & AlignTree & \textbf{0} & 4.47s & \textbf{0} & 2.82s & \textbf{0} & 4.84s & \textbf{0} & 3.95s \\ 
\midrule 
Qwen & RefusalClassifier & \textbf{0} & 23.1s & \textbf{0} & 30.35s & \textbf{0} & 17.73s & \textbf{0} & 54.87s \\ 
2.5-7B & SVMClassifier & \textbf{0} & 116.62s & \textbf{0} & 35.97s & \textbf{0} & 75.31s & \textbf{0} & 77.06s \\ 
 -Instruct & MultiRefusalsClassifier & \textbf{0} & \textbf{8.96}s & \textbf{0} & \textbf{13.96}s & \textbf{0} & \textbf{6.22}s & \textbf{0} & \textbf{27.54}s \\ 
 & AlignTreeLinear & \textbf{0} & 10.37s & \textbf{0} & 15.56s & \textbf{0} & 7.2s & \textbf{0} & 30.08s \\ 
 & AlignTree & \textbf{0} & 11.57s & \textbf{0} & 29.54s & \textbf{0} & 9.2s & \textbf{0} & 33.47s \\
\bottomrule
\end{tabular}
\end{small}
\caption{Comprehensive Refusal results for ablated defenses in AlignTree. The observed trend aligns with Table~\ref{tab:ablation_study}. }
\label{tab:full_refusal_ablation_study}
\end{table*}

\section{Refusal Classifier Discussion}

\label{appendix:refusal_classifier}
The high baseline ASR for Qwen models (Table~\ref{tab:qwen_asr}) compared to Llama models (Table~\ref{tab:llama_asr}) indicates weaker alignment in Qwen. This alignment quality dramatically affects RefusalClassifier performance: while achieving optimal results on Llama models, the classifier becomes nearly ineffective on Qwen models (Table~\ref{tab:full_asr_ablation_study}). This stark performance collapse—rather than gradual degradation—demonstrates that refusal detection in weakly aligned models presents fundamentally different interpretability challenges that render simple classification approaches impractical
\section{Use of existing assets}
All models used in this work are mentioned in Table~\ref {tab:model_used}, and datasets used are mentioned in Table~\ref {tab:datasets_used}. Please note this table is based on \citet{arditi2024refusallanguagemodelsmediated} collection of datasets.

\section{Compute statement}
\label{apdx:compute}
All experiments presented in this paper were run on a single NVIDIA-RTX-6000-Ada generation GPU with 48 GB of memory.

Training AlignTree, as described in Section~\ref{section:train_AlignTree}, takes approximately 3 minutes for the largest models; all defenses' execution times are reported in our results tables in Section~\ref{sec:full_results}.

\begin{table*}
\centering
\small
\begin{tabular}{l l l l}
\toprule
Model & Source & Accessed via & License \\
\midrule
\textsc{Qwen2.5 Instruct} & \citet{qwen2.5} & huggingface.co/Qwen & Tongyi Qianwen Research License\\
\textsc{Gemma3 IT}  &  \citet{gemmateam2025gemma3technicalreport} & huggingface.co/blog/gemma3 & Gemma Terms of Use \\
\textsc{Llama-3 Instruct} &  \citet{grattafiori2024llama3herdmodels} & huggingface.co/Meta-Llama-3-8B-Instruct & Meta Llama 3 Community License \\
\textsc{Llama Guard 3} &  \citet{meta2024llamaguard} & huggingface.co/Llama-Guard-3-1B & Meta Llama 3 Community License \\
\textsc{ChatGPT-4o} &  \citet{openai2024chatgpt4o} & chat.openai.com & OpenAI Terms of Use \\
\bottomrule
\end{tabular}
\caption{Models used in this work}
\label{tab:model_used}
\end{table*}

\begin{table*}[h!]
\centering
\small
\begin{tabular}{l l l l}
\toprule
Dataset & Source & Accessed via & License \\
\midrule
\textsc{AdvBench} & \citet{zou2023universaltransferableadversarialattacks} & github.com/llm-attacks/llm-attacks & MIT License \\
\textsc{TDC2023} & \citet{mazeika2024harmbenchstandardizedevaluationframework, mantas2023tdc} & github.com/centerforaisafety/tdc2023-starter-kit & MIT License \\
\textsc{HarmBench} & \citet{mazeika2024harmbenchstandardizedevaluationframework} & github.com/centerforaisafety/HarmBench & MIT License \\
\textsc{JailbreakBench} & \citet{patrick2024jailbreakbench} & github.com/JailbreakBench/jailbreakbench & MIT License \\
\textsc{MaliciousInstruct} & \citet{huang2023catastrophicjailbreakopensourcellms} & github.com/princeton-sysml/jailbreak\_llm & MIT License \\
\textsc{Alpaca} & \citet{alpaca} & huggingface.co/datasets/tatsu-lab/alpaca & Apache License 2.0 \\
\textsc{PIQA} & \citet{Bisk2020} & huggingface.co/datasets/ybisk/piqa & Apache License 2.0 \\
\textsc{ARC-Challenge} & \citet{clark2018thinksolvedquestionanswering} & huggingface.co/datasets/allenai/ai2\_arc & CC-BY-SA-4.0 \\
\textsc{OpenbookQA} & \citet{OpenBookQA2018} & huggingface.co/datasets/allenai/openbookqa & Apache License 2.0 \\
\textsc{SIQA} & \citet{siqa2019} & huggingface.co/datasets/allenai/social\_i\_qa & Apache License 2.0 \\
\textsc{JBShield (attacks)} & \citet{zhang2025jbshielddefendinglargelanguage} & github.com/NISPLab/JBShield & MIT License \\
\textsc{Garak} & \citet{derczynski2024garakframeworksecurityprobing} & github.com/NVIDIA/garak & Apache License 2.0 \\
\textsc{XSTest} & \citet{rottger-etal-2024-xstest} & huggingface.co/datasets/walledai/XSTest & CC-BY-4.0 \\
\bottomrule
\end{tabular}
\caption{Datasets used in this work}
\label{tab:datasets_used}
\end{table*}

\end{document}